\documentclass[letterpaper]{article} 
\usepackage[preprint]{aaai2027}  
\usepackage[hyphens]{url}  
\usepackage{graphicx} 
\usepackage{natbib}  
\usepackage{caption} 
\usepackage{algorithm}
\usepackage{algorithmic}

\usepackage{booktabs}
\usepackage{tabularx}
\usepackage{float}
\usepackage{amsmath}
\usepackage{amssymb}
\newtheorem{theorem}{Theorem}

\usepackage{newfloat}
\usepackage{listings}
\DeclareCaptionStyle{ruled}{labelfont=normalfont,labelsep=colon,strut=off} 
\floatstyle{ruled}
\newfloat{listing}{tb}{lst}{}
\floatname{listing}{Listing}

\usepackage{booktabs}

\title{Match One, Learn with Graph: One-to-Graph Query Collaboration with Backward Sharing for Object Detection}
\author {
    Wenxiao Fan\textsuperscript{\rm 1}\thanks{This work was conducted while Wenxiao Fan was an intern at JD.com.},
    Jingling Fu\textsuperscript{\rm 2}, 
    Luohang Liu\textsuperscript{\rm 2}, 
        Lichen Ma\textsuperscript{\rm 2,3}, 
    Yu He\textsuperscript{\rm 2}, 
    Zhiyang Yu\textsuperscript{\rm 2}, 
    Weishan Bi\textsuperscript{\rm 2}, 
    Junshi Huang\textsuperscript{\rm 2}, 
    Yan Li\textsuperscript{\rm 2}, 
    Gu Simiu\textsuperscript{\rm 2}, 
    Kan Li\textsuperscript{\rm 1}\corresponding
}
\affiliations {
    \textsuperscript{\rm 1}School of Computer Science, Beijing Institute of Technology\\
    \textsuperscript{\rm 2}JD.COM\\
    \textsuperscript{\rm 3}Institute of Artificial Intelligence and Robotics, Xi'an Jiaotong University\\
    \{wenxiaofan, likan\}@bit.edu.cn
}

\begin{document}

\maketitle

\begin{abstract}
One-to-one (O2O) matching enables Detection Transformers (DETRs) to perform end-to-end set prediction by assigning each object to a single positive query. However, the strongest classification, center, scale, and overlap evidence for an object is often distributed across multiple queries. This mismatch leaves only the matched owner positively supervised for the object, while other evidence-bearing queries receive no box target for it. We term this \emph{query knowledge fragmentation}. To exploit such complementary evidence without one-to-many supervision, we propose BS-O2G, a plug-in that builds a sparse prediction-aware graph from decoded features, boxes, and class distributions to organize query collaboration in feature and optimization spaces while preserving the original O2O matcher, positive labels, and objective. One-to-Graph (O2G) calibration propagates relative messages over this graph to consolidate query evidence in the forward pass, whereas Backward Sharing (BS) reuses its transposed detached adjacency to route gradients across persistent query basis vectors without changing the decoder input in the forward pass. Experiments across diverse DETR methods, backbones, COCO, and CrowdHuman show consistent gains and faster convergence with negligible parameter/FLOP growth and modest runtime overhead, supporting graph-based query collaboration as an alternative to expanding positive assignments.
\end{abstract}


\section{Introduction}
\label{sec:introduction}
Object detection aims to recognize all objects in an image. DETR~\cite{DBLP:conf/eccv/CarionMSUKZ20,hu2024dac} formulates this task as direct set prediction and uses one-to-one (O2O) Hungarian matching to remove hand-crafted anchors and non-maximum suppression. Subsequent variants substantially improve convergence and accuracy through sparse attention~\cite{DBLP:conf/iclr/ZhuSLLWD21}, denoising training~\cite{DBLP:conf/iclr/0097LL000NS23}, and stronger optimization strategies~\cite{DBLP:conf/cvpr/HuangLCYZS25}. Despite these advances, O2O supervision still designates only one matched query as the positive owner of each ground-truth object. A unique owner makes the training target unambiguous, but does not require the classification and localization criteria for an object to be best captured by the same query.

\begin{figure}[t]
    \centering
    \includegraphics[width=\linewidth]{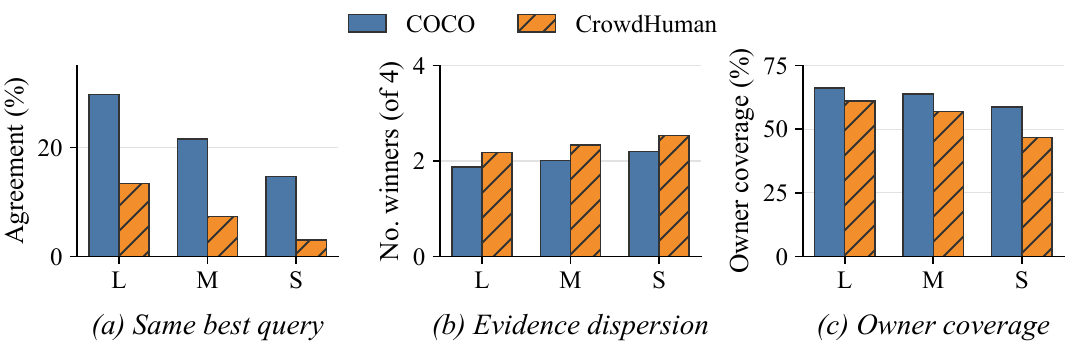}
    \caption{Query-evidence fragmentation on COCO and CrowdHuman under DEIM \cite{DBLP:conf/cvpr/HuangLCYZS25}. Panels show (a) agreement among the classification, center, scale, and IoU winners, (b) their number of distinct queries, and (c) matched-owner coverage across object sizes (L/M/S).
    }
    \label{fig:knowledge_fragmentation}
    \vspace{-1em}
\end{figure}

We examine this gap by decomposing each object's matching evidence into classification, center, scale, and IoU. As shown in Fig.~\ref{fig:knowledge_fragmentation}, both COCO and CrowdHuman exhibit a consistent scale-dependent trend: from large to small objects, winner agreement and matched-owner coverage decrease, whereas the number of distinct dimension-wise winners increases. Thus, for the same GT, classification, center, scale, and IoU evidence becomes increasingly dispersed across queries as object size decreases. This dispersion is most pronounced for small objects, whose four winners span $2.20$ queries with $58.65\%$ owner coverage on COCO and $2.53$ queries with $46.82\%$ coverage on CrowdHuman.
Thus, the limitation lies not merely in which query is selected by O2O matching, but in treating each query as a self-contained prediction unit rather than allowing prediction-related queries to collaborate over complementary evidence.
The matched owner may therefore capture only part of the available evidence, while non-owner evidence-bearing queries receive no box target for that object and are not positively supervised for it. We term this mismatch \emph{query knowledge fragmentation}.

Query knowledge fragmentation naturally motivates \emph{multi-query collaboration}: queries strong on complementary matching dimensions should be allowed to contribute jointly while only one retains positive ownership. A direct solution is one-to-many (O2M) training, which assigns the same object to multiple positive queries, either in the main assignment or through an auxiliary training branch~\cite{DBLP:conf/iccv/ZongS023}. However, O2M introduces extra training complexity and potential supervision conflicts with the primary O2O objective~\cite{DBLP:conf/cvpr/Zhao0WCDY024,DBLP:conf/cvpr/ZhangZH25}. We instead seek query collaboration at the feature and optimization levels, while keeping the matcher, positive labels, and detection objective strictly O2O.

To this end, we propose \textbf{BS-O2G}, which realizes prediction-aware query collaboration without introducing O2M supervision. BS-O2G maintains a persistent query basis and constructs a shared post-decoder graph from predicted features, boxes, and class distributions. On the forward path, One-to-Graph (O2G) calibration allows related normal queries to exchange information through relative graph messages. On the backward path, Backward Sharing (BS) reuses the transpose of the same detached affinity matrix so that their gradients jointly shape multiple persistent query basis vectors. Thus, O2G and BS coordinate collaboration in the feature and optimization spaces, respectively. BS is forward-identical, the original matcher and positive labels remain unchanged, and denoising queries are excluded from both components.
Designed as a lightweight plug-in for decoder-based DETR detectors, BS-O2G incurs only $0.69\%$ additional parameters and FLOPs on ResNet-50 and yields gains in several evaluated settings spanning multiple backbones and the COCO and CrowdHuman datasets.
Our contributions are three-fold:
\begin{itemize}
    \item We identify query knowledge fragmentation in O2O set prediction and show that dimension-wise evidence can be dispersed across multiple candidates even when only one query owns the positive target.
    \item We introduce BS-O2G to realize prediction-aware query collaboration with forward information exchange and backward basis-gradient sharing, without adding positive assignments.
    \item Experimental results demonstrate that BS-O2G improves detection performance across diverse settings, while introducing negligible parameter growth and only modest training overhead.
\end{itemize}
\FloatBarrier

\section{Related Work}
\label{sec:related_work}

\noindent\textbf{End-to-End Detection Transformers.}
DETR formulates object detection as direct set prediction and employs bipartite matching to produce non-redundant detections without non-maximum suppression~\cite{DBLP:conf/eccv/CarionMSUKZ20}. Subsequent studies improve its efficiency and optimization through sparse deformable attention~\cite{DBLP:conf/iclr/ZhuSLLWD21}, conditional spatial queries and dynamic anchor-box queries~\cite{DBLP:conf/iccv/MengCFZLYS021,DBLP:conf/iclr/LiuLZYQSZZ22}, and query denoising with stronger initialization~\cite{DBLP:conf/cvpr/LiZLGNZ22,DBLP:conf/iclr/0097LL000NS23}. More recent systems further refine distribution-based localization~\cite{DBLP:conf/iclr/PengLWZ0025}, combine efficient hybrid encoding with uncertainty-minimal query selection for real-time detection~\cite{DBLP:conf/cvpr/ZhaoLXWWDLC24}, or strengthen Dense O2O matching with matchability-aware optimization~\cite{DBLP:conf/cvpr/HuangLCYZS25}. Match-free supervision has also been explored as an alternative that removes explicit Hungarian assignment~\cite{DBLP:journals/corr/abs-2603-08514}. Despite these advances, the primary O2O objective used by most DETRs still assigns each ground-truth object to a unique positive query. Our work studies a complementary problem: how object-relevant evidence carried by related non-owner queries can participate in learning while the O2O assignment and detection objective remain unchanged.

\noindent\textbf{Relational Modeling among Object Queries.}
Relation Networks established appearance-and-geometry interaction among object candidates before the DETR era~\cite{DBLP:conf/cvpr/HuGZDW18}. Within DETRs, decoder self-attention enables implicit query interaction, motivating more explicit control of query cooperation and competition. Team DETR organizes query collaboration with position constraints~\cite{DBLP:conf/icip/QiuZXCFS23}; EASE-DETR encodes pairwise leading--trailing relations~\cite{DBLP:conf/cvpr/Gao0DZL24}; Relation-DETR injects dense relative-box geometry into self-attention~\cite{DBLP:conf/eccv/HouLZWCL24}; and LP-DETR and Dual-R-DETR progressively model spatial relations or route competitive and cooperative query pairs~\cite{DBLP:conf/icic/KangZDLZ25,official:icme/ZhangCHLK26}. MDS-DETR instead introduces confidence-guided causal masking in decoder self-attention to suppress duplicates under O2M supervision~\cite{DBLP:journals/corr/abs-2605-23507}. In contrast, O2G organizes prediction-aware query collaboration using a sparse post-decoder graph built jointly from final features, boxes, and class distributions; it propagates receiver-dependent relative messages and exposes the same detached relation to BS for backward sharing.

\noindent\textbf{Query Learning and Optimization.}
Query learning has been improved through dense-but-distinct candidate selection~\cite{DBLP:conf/cvpr/ZhangWWPLZLC23}, image-adaptive query generation~\cite{Kang_2026_CVPR}, parallel multi-time inquiries into image features~\cite{DBLP:conf/cvpr/NanLD025}, and better alignment between classification, localization, ranking, and matching~\cite{DBLP:conf/iccv/LiuRCZZLLHSZZ23,DBLP:conf/bmvc/CaiLWLG0024,DBLP:conf/nips/PuLHYYZHH23}. Training-time positive densification is realized through auxiliary O2M matching in H-DETR, group-wise assignment in Group DETR, auxiliary dense heads in Co-DETR, mixed supervision in MS-DETR, and multi-route supervision in Mr.~DETR~\cite{DBLP:conf/cvpr/JiaYHWYL00H23,DBLP:conf/iccv/ChenCWZYFHDZ023,DBLP:conf/iccv/ZongS023,DBLP:conf/cvpr/Zhao0WCDY024,DBLP:conf/cvpr/ZhangZH25}. OD-DETR and query-selection distillation instead stabilize query learning through teacher guidance~\cite{DBLP:conf/ijcai/WuSL24,DBLP:journals/corr/abs-2409-06443}. BS operates on a different axis: it augments image-conditioned query content with a persistent basis and routes existing O2O gradients among graph-related basis vectors, enabling optimization-space collaboration without additional positive assignments, teachers, or auxiliary prediction routes.

\section{Method}
\label{sec:method}





\begin{figure*}[t]
    \centering
    \includegraphics[width=\textwidth]{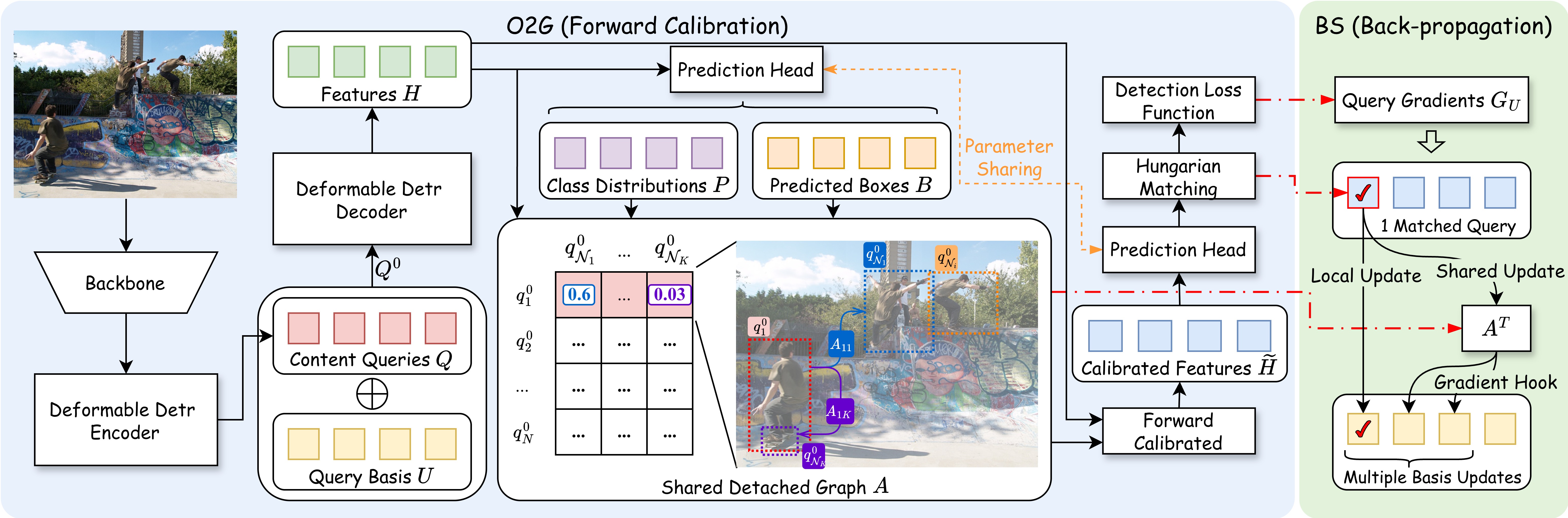}
    \caption{Overview of BS-O2G. Image-conditioned content queries $Q$ are augmented with the persistent query basis $U$ and decoded into $H$, whose features and predictions define a detached graph $A$. O2G uses $A$ for forward feature calibration, while BS uses $A^\top$ to share query-basis gradients across graph neighbors without changing O2O supervision.}
    \label{fig:bqs_o2g_framework}
    \vspace{-1em}
\end{figure*}

Motivated by query knowledge fragmentation, BS-O2G constructs a shared post-decoder graph from hidden states and predictions to realize collaboration without O2M supervision. O2G lets each query read graph neighbors through $A$ for feature-space collaboration, whereas Backward Sharing (BS) routes query-basis gradients along $A^\top$ for optimization-space collaboration. Hungarian matching, positive assignment, and the detection objective remain strictly O2O throughout.
Fig.~\ref{fig:bqs_o2g_framework} summarizes the complete framework. 

\noindent\textbf{Persistent Query Basis.}
\label{sec:basis}
For an image $x$, a two-stage detector selects $N$ image-conditioned content queries $Q(x)=[q_1(x),\ldots,q_N(x)]^\top\in\mathbb{R}^{N\times d}$ and reference boxes $R(x)$ from encoder proposals \cite{DBLP:conf/iclr/ZhuSLLWD21,DBLP:conf/cvpr/HuangLCYZS25}. Because the proposal occupying rank $i$ and its content vary across images, $q_i(x)$ provides no persistent learnable state tied to query index $i$. We therefore augment each content query with a learnable persistent query basis, implemented as a bank of query vectors, $U=[u_1,\ldots,u_N]^\top\in\mathbb{R}^{N\times d}$ shared across images:
\begin{equation}
q_i^0(x)=q_i(x)+u_i
\label{eq:query_decomposition}
\end{equation}
Here $q_i(x)$ carries image-specific visual evidence, whereas $u_i$ provides a persistent content-side carrier on which BS accumulates and shares optimization signals across images. Their sum forms the decoder content query. The reference-box and positional-query paths remain unchanged; $U$ is neither a reference box nor a positional query.

\noindent\textbf{Shared Prediction-Aware Graph.}
To organize collaboration around complementary evidence, we construct a sparse detection-specific graph from the final decoder states and preliminary predictions of normal queries. Let $H=[h_1,\ldots,h_N]^\top$ denote the decoder features obtained from $Q(x)+U$. Applying the prediction heads to $H$ yields preliminary boxes $b_i$, logits $z_i$, and class probabilities $p_i=\sigma(z_i)$. For each pair $i\neq j$, we define the affinity
\begin{equation}
s_{ij}=\frac{1}{\sqrt d}\operatorname{cos}(h_i,h_j)
+\operatorname{IoU}(b_i,b_j)
+\operatorname{cos}(p_i,p_j).
\label{eq:graph_affinity}
\end{equation}
The three terms capture feature, localization, and class consistency, respectively. To avoid message passing over a fully connected query graph, we sparsify it through row-wise top-$K$ selection. For each target query $i$, we rank all other queries $j\neq i$ by $s_{ij}$ and retain the $K$ highest-scoring queries as its neighborhood $\mathcal{N}_i$. Thus, $K$ controls the neighborhood size and graph sparsity, restricting subsequent graph operations to $NK$ directed edges instead of $N(N-1)$. We then normalize the selected affinities within each neighborhood:
\begin{equation}
A_{ij}=\frac{\mathbb{I}[j\in\mathcal{N}_i]\exp(s_{ij}/\tau)}
{\sum_{k\in\mathcal{N}_i}\exp(s_{ik}/\tau)}.
\label{eq:graph_adjacency}
\end{equation}
This produces a directed sparse adjacency $A$, where $A_{ij}$ quantifies how strongly query $i$ relates to query $j$ under the current prediction state. Because the top-$K$ neighborhood is selected independently for each query, $j\in\mathcal{N}_i$ does not necessarily imply $i\in\mathcal{N}_j$; hence, $A$ is generally asymmetric. We detach $A$ and reuse it for O2G forward message passing and BS backward gradient routing. Denoising queries are excluded from the graph and both operations.

\noindent\textbf{O2G Forward Calibration.}
\label{sec:o2g}
O2G realizes feature-space collaboration by consolidating complementary query evidence. It first encodes the relative state of each directed edge $i\leftarrow j$ as
\begin{equation}
v_{ij}=\phi_m([h_j-h_i;\rho(b_i,b_j);\operatorname{IoU}(b_i,b_j);p_j-p_i]).
\label{eq:o2g_message}
\end{equation}
Here, $v_{ij}$ denotes the relative message sent from neighbor query $j$ to target query $i$, $\phi_m$ is a learnable two-layer MLP, and $\rho(b_i,b_j)=[(c_j-c_i)/s_i;\log(s_j/s_i)]$, with $c_i$ and $s_i$ denoting the center and size of $b_i$, respectively.

O2G then aggregates the edge messages to calibrate the target query and obtains the final predictions:
\begin{equation}
\begin{aligned}
\widetilde h_i&=h_i+\gamma W_o\sum_{j\in\mathcal{N}_i}A_{ij}v_{ij},\\
\widehat z_i&=f_{\mathrm{cls}}(\widetilde h_i),\qquad
\widehat b_i=f_{\mathrm{box}}(\widetilde h_i).
\end{aligned}
\label{eq:o2g_update}
\end{equation}
Here, $W_o$ maps the aggregated relational message back to the decoder feature space, while the learnable scalar $\gamma$ controls the strength of the O2G relational residual. The update preserves the target's decoded representation while incorporating relation-aware cues from its $K$ neighbors. The calibrated features $\widetilde H$ are then fed to the classification and box heads, producing $\widehat Z=[\widehat z_i]_{i=1}^{N}$ and $\widehat B=[\widehat b_i]_{i=1}^{N}$ for Hungarian matching and the detection objective. We denote the initialization of this gate by $\gamma_0$ and set $\gamma_0=0$, so that training starts from the uncalibrated decoder output. O2G is used in both training and inference.

\begin{table*}[t]
\centering
\small
\setlength{\tabcolsep}{2.5pt}
\begin{tabular}{@{}l|cc|cccccc@{}}
\toprule
\textbf{Method}
& \textbf{Epochs}
& \textbf{Queries}
& \textbf{AP}
& $\mathbf{AP}_{50}$
& $\mathbf{AP}_{75}$
& $\mathbf{AP}_{S}$
& $\mathbf{AP}_{M}$
& $\mathbf{AP}_{L}$ \\
\midrule
DDQ-DETR~\cite{DBLP:conf/cvpr/ZhangWWPLZLC23}
& 24 & 900
& 52.0 & 69.5 & 57.2 & 35.2 & 54.9 & 65.9 \\
Stable-DINO~\cite{DBLP:conf/iccv/LiuRCZZLLHSZZ23}
& 24 & 900
& 51.5 & 68.5 & 56.3 & 35.2 & 54.7 & 66.5 \\
MS-DETR~\cite{DBLP:conf/cvpr/Zhao0WCDY024}
& 24 & 900
& 51.7 & 68.7 & 56.5 & 34.0 & 55.4 & 65.5 \\
MR-DETR~\cite{DBLP:conf/cvpr/ZhangZH25}
& 24 & 900
& 51.4 & 69.0 & 56.2 & 34.9 & 54.8 & 66.0 \\
Relation-DETR~\cite{DBLP:conf/eccv/HouLZWCL24}
& 24 & 900
& 52.1 & 69.7 & 56.6 & 36.1 & 56.0 & 66.5 \\
LP-DETR~\cite{DBLP:conf/icic/KangZDLZ25}
& 24 & --
& 52.5 & 70.0 & 57.2 & 36.2 & 56.3 & 67.1 \\
PaQ-DINO~\cite{Kang_2026_CVPR}
& 24 & 900
& 52.6 & 69.7 & 56.9 & 35.7 & 56.4 & 67.0 \\
DiffuAlignDETR~\cite{official:iclr/NawarBT26}
& 24 & 900
& 51.9 & 69.2 & 56.4 & 34.9 & 55.6 & 66.2 \\
DEIM $^{++}$ \cite{DBLP:conf/cvpr/HuangLCYZS25}
& 24 & 300
& $53.0_{\scriptscriptstyle\pm0.07}$
& $70.9_{\scriptscriptstyle\pm0.07}$
& $57.7_{\scriptscriptstyle\pm0.18}$
& $36.0_{\scriptscriptstyle\pm0.26}$
& $57.7_{\scriptscriptstyle\pm0.07}$
& $70.1_{\scriptscriptstyle\pm0.33}$ \\
\textbf{BS-O2G (ours)}
& 24 & 300
& $\mathbf{53.5}_{\scriptscriptstyle\pm0.09}$
& $\mathbf{71.3}_{\scriptscriptstyle\pm0.09}$
& $\mathbf{58.1}_{\scriptscriptstyle\pm0.19}$
& $\mathbf{36.9}_{\scriptscriptstyle\pm0.29}$
& $\mathbf{58.0}_{\scriptscriptstyle\pm0.24}$
& $\mathbf{70.5}_{\scriptscriptstyle\pm0.20}$ \\
\midrule
DiffuDETR~\cite{official:iclr/NawarBT26}
& 50 & 300
& 50.2 & 66.8 & 55.2 & 33.3 & 53.9 & 65.8 \\
DiffuDINO~\cite{official:iclr/NawarBT26}
& 50 & 900
& 51.9 & 69.4 & 55.7 & 35.8 & 55.7 & 67.1 \\
RT-DETRv2 $^{++}$ \cite{DBLP:journals/corr/abs-2407-17140}
& 72 & 300
& $53.2_{\scriptscriptstyle\pm0.12}$
& $71.3_{\scriptscriptstyle\pm0.09}$
& $57.5_{\scriptscriptstyle\pm0.21}$
& $36.1_{\scriptscriptstyle\pm0.22}$
& $57.5_{\scriptscriptstyle\pm0.07}$
& $70.7_{\scriptscriptstyle\pm0.18}$ \\
DEIM $^{++}$ \cite{DBLP:conf/cvpr/HuangLCYZS25}
& 60 & 300
& $54.0_{\scriptscriptstyle\pm0.21}$
& $71.8_{\scriptscriptstyle\pm0.10}$
& $58.7_{\scriptscriptstyle\pm0.19}$
& $36.9_{\scriptscriptstyle\pm0.12}$
& $58.5_{\scriptscriptstyle\pm0.16}$
& $70.8_{\scriptscriptstyle\pm0.34}$ \\
\textbf{BS-O2G (ours)}
& 36 & 300
& $54.1_{\scriptscriptstyle\pm0.18}$
& $72.0_{\scriptscriptstyle\pm0.12}$
& $58.8_{\scriptscriptstyle\pm0.38}$
& $36.8_{\scriptscriptstyle\pm0.24}$
& $58.5_{\scriptscriptstyle\pm0.37}$
& $71.1_{\scriptscriptstyle\pm0.18}$ \\
\textbf{BS-O2G (ours)}
& 60 & 300
& $\mathbf{54.5}_{\scriptscriptstyle\pm0.05}$
& $\mathbf{72.5}_{\scriptscriptstyle\pm0.13}$
& $\mathbf{59.2}_{\scriptscriptstyle\pm0.19}$
& $\mathbf{37.8}_{\scriptscriptstyle\pm0.30}$
& $\mathbf{58.7}_{\scriptscriptstyle\pm0.31}$
& $\mathbf{71.3}_{\scriptscriptstyle\pm0.22}$ \\
\bottomrule
\end{tabular}
\caption{Comparison with state-of-the-art object detectors on COCO val2017 using a ResNet-50 backbone. The best results within the 24-epoch and long-schedule comparison blocks are shown in bold. The superscript $^{++}$ denotes we re-implement the methods and report the corresponding results with standard deviations.}
\label{tab:main_results}
\vspace{-1.2em}
\end{table*}

\noindent\textbf{Graph-Weighted Backward Sharing.}
\label{sec:bqs}
While O2G coordinates feature-space collaboration in the forward path, BS reuses the detached graph to coordinate optimization signals among persistent query basis vectors. In practice, this redistribution is applied through a backward hook after $A$ is constructed; equivalently, its Jacobian can be expressed by the following straight-through form. Defining $\Delta U=AU-U$, we write
\begin{equation}
\begin{aligned}
U^{\mathrm{BS}}&=U+\lambda_{\mathrm B}\Delta U
-\operatorname{sg}\!\left(\lambda_{\mathrm B}\Delta U\right),\\
Q_{\mathrm{BS}}^0(x)&=Q(x)+U^{\mathrm{BS}}
\overset{\mathrm{forward}}{=}Q(x)+U.
\end{aligned}
\label{eq:bqs_forward_identity}
\end{equation}
Here, $\operatorname{sg}(\cdot)$ denotes stop-gradient, and $\lambda_{\mathrm B}$ controls the BS gradient-sharing strength, interpolating between the local update and graph-routed sharing in Eq.~\eqref{eq:bqs_gradient}. The residual and its stop-gradient copy have identical forward values and cancel numerically, whereas only the former contributes derivatives. BS therefore leaves the decoder input unchanged while introducing a graph-dependent backward path.

Let $G_U(x)=\partial\mathcal{L}(x)/\partial U^{\mathrm{BS}}$ denote the per-image gradient arriving at the straight-through output. Since $A$ is detached, differentiating Eq.~\eqref{eq:bqs_forward_identity} gives
\begin{equation}
\begin{aligned}
G_U^{\mathrm{BS}}(x)
=\underbrace{(1-\lambda_{\mathrm B})G_U(x)}_{\text{Local Update}}
+\underbrace{\lambda_{\mathrm B}A^\top G_U(x)}_{\text{Shared Update}}
\end{aligned}
\label{eq:bqs_gradient}
\end{equation}
Eq.~\eqref{eq:bqs_gradient} decomposes the basis gradient into two components. The \emph{Local Update} retains each query's gradient at its own basis vector, while the \emph{Shared Update} routes gradients through $A^\top$. Since each row of $A$ has only $K$ nonzero entries, $A^\top$ performs sparse one-hop reverse-edge routing rather than global averaging. If $A_{ij}$ allows query $i$ to read query $j$ in O2G, a fraction $\lambda_{\mathrm B}A_{ij}$ of query $i$'s gradient is routed to $u_j$. Applied to all normal queries, BS redistributes existing detection-loss gradients among graph-related persistent basis vectors without adding query--GT assignments or auxiliary losses or changing the O2O objective. One query gradient can therefore jointly update multiple graph-neighbor basis vectors without creating O2M positive supervision. Equation~\eqref{eq:bqs_gradient} is the clean-region form in which the implementation's numerical gradient and basis sanitizers are inactive; App.~\ref{app:bqs_derivation} states the exact sanitized implementation and full Jacobian derivation.

We use warm-up for $\lambda_{\mathrm B}$, keeping it at zero initially and then increasing it linearly to the target value. At inference, gradient routing is absent; the learned basis $U$ is retained, and O2G continues to perform forward graph calibration.

\noindent\textbf{Objective.}
\label{sec:objective}
The matcher and training objective are unchanged:
\begin{equation}
\mathcal{L}_{\mathrm{BS\text{-}O2G}}
=\mathcal{L}(\widehat B,\widehat Z;Y).
\label{eq:objective}
\end{equation}
BS adds no auxiliary loss, and O2G modifies only the final normal-query prediction. BS-O2G therefore coordinates query collaboration in feature and optimization spaces while leaving the decoder architecture, matcher, and detection objective unchanged, making it a plug-in for decoder-based DETR-style detectors.
The complete training and inference procedure is summarized in Algorithm~\ref{alg:bqs_o2g_complete} (App.~\ref{app:bqs_o2g_algorithm}).

\noindent\textbf{Simplified Analysis of BS-O2G.}
We summarize the two effects central to BS-O2G under simplified local models aligned with the shared graph $A$. Complete proofs are provided in Apps.~\ref{app:proof_o2g_mse} and~\ref{app:proof_bqs_contraction}.

\noindent\textbf{O2G reduces representation error.}
Consider the normalized feature-difference slice
\begin{equation}
\widetilde h_i^{\mathrm{lin}}
=h_i+\gamma\sum_{j\in\mathcal N_i}A_{ij}(h_j-h_i),
\label{eq:o2g_linear_slice}
\end{equation}
Let $\nu^2$ denote the per-coordinate feature-noise variance, $K_i^{\mathrm{eff}}=(\sum_{j\in\mathcal N_i}A_{ij}^2)^{-1}$ the effective neighborhood size, and $\bar\delta_i$ the affinity-weighted neighbor bias.

\begin{theorem}[O2G error reduction]
\label{thm:o2g_mse}
Under the conditional model in App.~\ref{app:proof_o2g_mse}, Eq.~\eqref{eq:o2g_linear_slice} has lower representation MSE than the uncalibrated feature if and only if
\begin{equation}
0<\gamma<\frac{2d\nu^2}
{d\nu^2(1+1/K_i^{\mathrm{eff}})+\|\bar\delta_i\|_2^2}.
\label{eq:o2g_improvement_interval}
\end{equation}
\end{theorem}

Thus, O2G suppresses query-specific noise when its neighbors are sufficiently coherent; larger $K_i^{\mathrm{eff}}$ reduces variance, while $\|\bar\delta_i\|_2^2$ quantifies the cost of impure neighbors.

\noindent\textbf{BS performs graph low-pass filtering in optimization space.}
To expose its spectral effect, consider the balanced case in which the detached adjacency $A$ is symmetric and connected, with eigenvalues $1=\mu_1>\mu_2\ge\cdots\ge\mu_N\ge-1$. Let $P_{\mathrm B}=(1-\lambda_{\mathrm B})I_N+\lambda_{\mathrm B}A^\top$, so that $G_U^{\mathrm{BS}}=P_{\mathrm B}G_U$, and decompose $G_U$ into its mean $\bar g_U=N^{-1}\mathbf{1}_N^\top G_U$ and disagreement $G_{U,\perp}=G_U-\mathbf{1}_N\bar g_U$, where $\mathbf{1}_N\in\mathbb{R}^N$ is the all-ones vector.

\begin{theorem}[BS graph low-pass filtering]
\label{thm:bqs_contraction}
If $0<\lambda_{\mathrm B}\le1/2$, then BS preserves the mean gradient and contracts the graph non-consensus component:
\begin{equation}
\frac{1}{N}\mathbf{1}_N^\top G_U^{\mathrm{BS}}=\bar g_U,
\qquad
\|P_{\mathrm B}G_{U,\perp}\|_F
\le\rho_{\mathrm B}\|G_{U,\perp}\|_F,
\label{eq:bqs_contraction}
\end{equation}
where $\rho_{\mathrm B}=1-\lambda_{\mathrm B}(1-\mu_2)<1$.
\end{theorem}

The response of graph eigenmode $r$ is $p_r=1-\lambda_{\mathrm B}(1-\mu_r)$, preserving consensus ($p_1=1$) while increasingly attenuating modes with smaller $\mu_r$. Thus, BS smooths query-specific gradient disagreement over the affinity graph in optimization space. This theorem characterizes the balanced limit of the directed top-$K$ graph used in practice.


\begin{table}[t]
\centering
\small
\setlength{\tabcolsep}{2.0pt}
\resizebox{\columnwidth}{!}{%
\begin{tabular}{@{}lllcccc@{}}
\toprule
\textbf{Base Detector}
& \textbf{Backbone}
& \textbf{Config.}
& \textbf{AP}
& $\mathbf{AP}_{50}$
& $\mathbf{AP}_{75}$
& $\mathbf{AP}_{M}$ \\
\midrule
DINO & ResNet-50 & Baseline
& 48.5 & 65.9 & 53.0 & \textbf{52.0} \\
& & \textbf{+ BS-O2G}
& \textbf{48.8} & \textbf{66.3} & \textbf{53.4} & 51.9 \\
\midrule
RT-DETRv2 & ResNet-50 & Baseline
& 53.2 & 71.3 & 57.5 & 57.5 \\
& & \textbf{+ BS-O2G}
& \textbf{53.6} & \textbf{71.7} & \textbf{58.1} & \textbf{58.2} \\
\midrule
RT-DETRv2 & ResNet-101 & DEIM
& 55.5 & 73.5 & \textbf{60.5} & 60.1 \\
& & \textbf{+ BS-O2G}
& \textbf{55.8} & \textbf{74.0} & 60.3 & \textbf{60.5} \\
\midrule
D-FINE-L & HGNetV2-B4 & DEIM
& 54.2 & 71.9 & \textbf{59.1} & 58.7 \\
& & \textbf{+ BS-O2G}
& \textbf{54.5} & \textbf{72.2} & \textbf{59.1} & \textbf{59.3} \\
\midrule
D-FINE-X & HGNetV2-B5 & DEIM
& 56.2 & 73.8 & 61.1 & 61.4 \\
& & \textbf{+ BS-O2G}
& \textbf{56.6} & \textbf{74.2} & \textbf{61.4} & \textbf{61.6} \\
\bottomrule
\end{tabular}
}
\caption{Plug-in evaluation of BS-O2G across diverse detector--backbone configurations on COCO val2017.
}
\label{tab:backbone_results}
\vspace{-1em}
\end{table}

\section{Experiments}
\label{sec:experiments}


\noindent\textbf{Datasets and protocol.}
We conduct experiments on MS-COCO dataset \cite{DBLP:conf/eccv/LinMBHPRDZ14} and CrowdHuman \cite{DBLP:journals/corr/abs-1805-00123}. 
The MS-COCO and CrowdHuman experiments follow the protocol of DEIM~\cite{DBLP:conf/cvpr/HuangLCYZS25}. Our default model is DEIM-RT-DETRv2~\cite{DBLP:conf/cvpr/HuangLCYZS25} with a ResNet-50 backbone~\cite{DBLP:conf/cvpr/HeZRS16}, while experiments on other detectors and backbones retain their original training recipes. Unless otherwise specified, reproduced results use three seeds, $\{0,42,1027\}$; single-valued entries are means, while mean$\pm$standard deviation entries report both statistics. 
All reproduced models are trained on 8 NVIDIA H200 GPUs.

\noindent\textbf{Training and method settings.}
For the default COCO experiments, models are evaluated at $640\times640$ and optimized using AdamW with a learning rate of $2\times10^{-4}$, a backbone learning rate of $2\times10^{-5}$, weight decay of $10^{-4}$, and a total batch size of 16. 
BS-O2G uses $N=300$, $d=256$, $K=8$, and $\tau=0.7$. We initialize $U\sim\mathcal{N}(0,0.02^2)$ and $\gamma_0=0$. BS starts at epoch 8 and linearly increases $\lambda_{\mathrm B}$ to $0.02$ over six epochs. O2G is used during training and inference, whereas BS is training-only; the original Hungarian matcher and detection objective remain unchanged. Additional implementation details are provided in App.~\ref{app:implementation_details}.

\begin{table}[htbp]
\centering
\small
\setlength{\tabcolsep}{2.8pt}
\resizebox{\columnwidth}{!}{%
\begin{tabular}{@{}lccccccc@{}}
\toprule
\textbf{Method}
& \textbf{Epochs}
& \textbf{AP}
& $\mathbf{AP}_{50}$
& $\mathbf{AP}_{75}$
& $\mathbf{AP}_{S}$
& $\mathbf{AP}_{M}$
& $\mathbf{AP}_{L}$ \\
\midrule
D-FINE-L
& 120 & 56.0 & 87.2 & 59.4 & 29.0 & 46.1 & 54.6 \\
\quad + DEIM
& 120 & 57.5 & \textbf{87.6} & 62.9
& \textbf{33.2} & 48.7 & 55.7 \\
\textbf{\quad + BS-O2G}
& 90 & 58.0 & 86.6 & 63.6 & 30.6 & 50.2 & 68.4 \\
\textbf{\quad + BS-O2G}
& 120 & \textbf{58.9} & 87.2 & \textbf{64.7}
& 31.7 & \textbf{51.1} & \textbf{69.3} \\
\bottomrule
\end{tabular}
}
\caption{Comparison on CrowdHuman. The D-FINE-L and DEIM results are taken from~\cite{DBLP:conf/cvpr/HuangLCYZS25}.
}
\label{tab:crowdhuman_deim_reference}
\vspace{-0.6em}
\end{table}

\begin{table}[t]
\centering
\small
\begin{tabular}{@{}lcccc@{}}
\toprule
\textbf{Method}
& \textbf{Params}
& \textbf{FLOPs}
& \textbf{Latency (ms)}
& \textbf{FPS} \\
\midrule
DEIM
& 42.94
& 133.50
& $18.54{\pm}0.39$
& 53.9 \\
\quad + O2G
& 43.16
& 134.42
& $20.18{\pm}0.35$
& 49.6 \\
\textbf{\quad + BS-O2G}
& 43.24
& 134.42
& $20.37{\pm}0.32$
& 49.1 \\
\bottomrule
\end{tabular}
\caption{Model complexity and inference efficiency with a ResNet-50 backbone. Parameters are reported in millions and FLOPs in GFLOPs. Latency and FPS are measured with batch size 1 using $640\times640$ inputs and AMP.
}
\label{tab:efficiency}
\vspace{-1em}
\end{table}


\subsection{Effectiveness of BS-O2G}

\noindent\textbf{Comparison with ResNet-50 DETRs.}
Tab.~\ref{tab:main_results} compares BS-O2G with recent DETR variants on COCO \texttt{val2017}, separating the standard 24-epoch setting from longer training schedules. Notably, BS-O2G uses only 300 queries, one-third of the 900-query budget adopted by most competitors in the 24-epoch block, yet achieves the best result of $53.5$ AP. It outperforms DEIM~\cite{DBLP:conf/cvpr/HuangLCYZS25} by $0.5$ AP and PaQ-DINO~\cite{Kang_2026_CVPR}, the strongest 900-query competitor, by $0.9$ AP. Thus, the improvement is obtained without expanding the decoder query budget. Relative to DEIM, the gains are consistent across all COCO metrics: $+0.4$ AP$_{50}$, $+0.4$ AP$_{75}$, $+0.9$ AP$_S$, $+0.3$ AP$_M$, and $+0.4$ AP$_L$. The three-seed result also varies little.

The advantage persists under longer training while retaining the same budget. BS-O2G reaches $54.1$ AP at 36 epochs, matching the $54.0$ AP of DEIM trained for 60 epochs. Extending BS-O2G to 60 epochs further improves performance to $54.5$ AP, surpassing the corresponding DEIM baseline by $0.5$ AP. The improvement remains consistent across object scales, including $+0.9$ AP$_S$, $+0.2$ AP$_M$, and $+0.5$ AP$_L$, showing that BS-O2G improves both training efficiency and the final accuracy.

\begin{table}[t]
\centering
\footnotesize
\begin{tabular}{@{}cl|cccc@{}}
\toprule
\textbf{Epoch}
& \textbf{Method}
& $\mathbf{AR}_{10}$
& $\mathbf{AR}_{100}$
& $\mathbf{AR}_{S}$
& $\mathbf{AR}_{M}$ \\
\midrule
24 & DEIM
& 65.98 & 72.56 & 56.03 & 76.91 \\
24 & \textbf{+ BS-O2G}
& \textbf{66.57} & \textbf{73.10} & \textbf{57.14} & \textbf{77.36} \\
\midrule
36 & DEIM
& 66.45 & 73.05 & 55.37 & 77.37 \\
36 & \textbf{+ BS-O2G}
& \textbf{66.70} & \textbf{73.30} & \textbf{56.49} & \textbf{77.51} \\
\midrule
60 & DEIM
& 66.98 & 73.42 & 56.22 & \textbf{77.58} \\
60 & \textbf{+ BS-O2G}
& \textbf{67.29} & \textbf{73.62} & \textbf{58.26} & 77.45 \\
\midrule
\multicolumn{2}{l|}{\textbf{Avg. $\Delta$}}
& \textbf{+0.39} & \textbf{+0.33} & \textbf{+1.43} & \textbf{+0.15} \\
\bottomrule
\end{tabular}
\caption{Average recall comparison between BS-O2G and DEIM on COCO val2017 using ResNet-50 backbone.
}
\label{tab:ar_results}
\vspace{-0.6em}
\end{table}

\begin{table}[htbp]
\centering
\small
\begin{tabular}{@{}lccc@{}}
\toprule
\textbf{Set-level error ($\downarrow$)}
& \textbf{DEIM}
& \textbf{O2G}
& \textbf{BS-O2G} \\
\midrule
Duplicate FP & 3060 & 2794\,\raisebox{-0.45ex}{\scriptsize($-8.7\%$)} & \textbf{2710}\,\raisebox{-0.45ex}{\scriptsize($-11.4\%$)} \\
Localization FP & 8139 & 8047\,\raisebox{-0.45ex}{\scriptsize($-1.1\%$)} & \textbf{7944}\,\raisebox{-0.45ex}{\scriptsize($-2.4\%$)} \\
\bottomrule
\end{tabular}
\caption{Set-level false-positive counts on COCO val.
}
\label{tab:query_set_competition}
\vspace{-1em}
\end{table}

\noindent\textbf{Generalization across detectors and backbones.}
Tab.~\ref{tab:backbone_results} evaluates BS-O2G as a plug-in across DINO~\cite{DBLP:conf/iclr/0097LL000NS23}, RT-DETRv2~\cite{DBLP:journals/corr/abs-2407-17140}, and D-FINE~\cite{DBLP:conf/iclr/PengLWZ0025} with ResNet-50/101~\cite{DBLP:conf/cvpr/HeZRS16} and HGNetV2-B4/B5~\cite{DBLP:journals/corr/abs-2103-05959} backbones. BS-O2G improves AP for all five detector--backbone configurations by $0.3$--$0.4$ points and raises AP$_{50}$ by $0.3$--$0.5$ points. The gains are particularly clear for RT-DETRv2 with ResNet-50, where AP$_{75}$ and AP$_M$ increase by $0.6$ and $0.7$ points, respectively, and for D-FINE-L, where AP$_M$ improves by $0.6$ points. On the strongest D-FINE-X configuration, BS-O2G further increases AP from $56.2$ to $56.6$. Although individual metrics show small fluctuations, including $-0.1$ AP$_M$ for DINO and $-0.2$ AP$_{75}$ for RT-DETRv2 with ResNet-101, the consistent AP gains demonstrate that BS-O2G transfers beyond the default setting to diverse detector architectures and backbones.

\noindent\textbf{Results on crowded scenes.}
Tab.~\ref{tab:crowdhuman_deim_reference} evaluates BS-O2G on the challenging CrowdHuman benchmark. With only 90 training epochs, BS-O2G reaches $58.0$ AP, already surpassing the $57.5$ AP of DEIM trained for 120 epochs while using $25\%$ fewer epochs. Extending training to 120 epochs further improves BS-O2G to $58.9$ AP, outperforming DEIM and the original D-FINE-L by $1.4$ and $2.9$ points, respectively. These results show that BS-O2G transfers effectively to dense pedestrian detection and provides clear overall gains under both shorter and matched training schedules.

\noindent\textbf{Computational Efficiency.}
\label{sec:efficiency}
Tab.~\ref{tab:efficiency} compares the model complexity and inference efficiency of the DEIM baseline, O2G, and the complete BS-O2G model. BS-O2G increases both parameters and GFLOPs by only $0.69\%$ over the baseline. Its batch-1 latency increases from $18.54$ to $20.37$ ms, while retaining $49.1$ FPS. O2G and BS-O2G have identical GFLOPs and differ by only $0.18$ ms in measured latency because graph-based BS routing is training-only.




\subsection{Does BS-O2G Improve Fragmentation Outcomes?}

\noindent\textbf{Average recall.}
Tab.~\ref{tab:ar_results} further compares the recall performance of BS-O2G and DEIM. Across the 24-, 36-, and 60-epoch schedules, BS-O2G consistently improves both $\mathrm{AR}_{10}$ and $\mathrm{AR}_{100}$ over DEIM, with average gains of $+0.39$ and $+0.33$, respectively. The largest improvement occurs for small objects: $\mathrm{AR}_{S}$ increases by $1.43$ points on average and by $2.04$ points at 60 epochs.
This size-dependent pattern is consistent with the fragmentation motivation in Fig.~\ref{fig:knowledge_fragmentation}, where small objects exhibit stronger evidence dispersion and lower matched-owner coverage. Under the same prediction budget, the higher AR indicates that BS-O2G recovers more ground-truth objects, especially at the small scale.


\begin{figure}[t]
\centering
\includegraphics[width=\linewidth]{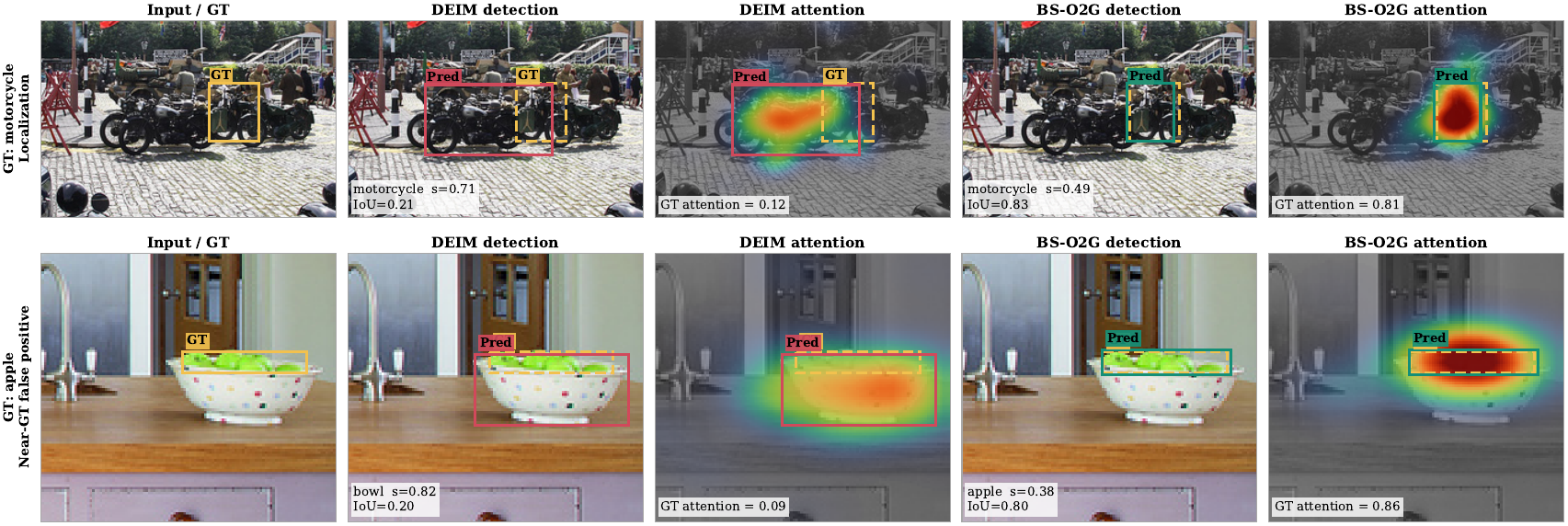}
\caption{
    Case studies of BS-O2G. Yellow dashed boxes indicate GT, while red and green boxes indicate DEIM and BS-O2G predictions, respectively. 
    Additional cases are shown in App.~\ref{app:paired_attention_cases}.
    Zoom in to see details.
}
\label{fig:paired_attention_main}
\vspace{-0.4em}
\end{figure}

\begin{table}[htbp]
\centering
\small
\begin{tabular}{@{}lcc@{}}
\toprule
\textbf{Query-family metric ($\uparrow$)}
& \textbf{DEIM}
& \textbf{BS-O2G} \\
\midrule
True-class family queries (\%) & 51.9 & \textbf{67.2} \\
True-class edge mass (\%) & 51.2 & \textbf{68.1} \\
Neighbor IoU to GT (\%) & 27.0 & \textbf{34.3} \\
\bottomrule
\end{tabular}
\caption{Query-family quality for targets missed by DEIM but recovered by BS-O2G.}
\label{tab:query_family_quality}
\vspace{-1em}
\end{table}

\noindent\textbf{BS-O2G reshapes the final query-set output.}
We audit the final predictions over the full validation set. A \emph{duplicate FP} is an extra prediction for an object that has already been correctly detected, whereas a \emph{localization FP} predicts the correct class near a target but fails to reach the TP IoU threshold. As shown in Tab.~\ref{tab:query_set_competition}, O2G and BS-O2G reduce both errors; the full method lowers duplicate FP by $11.4\%$ and localization FP by $2.4\%$ relative to DEIM. This shows that our method changes the set-level competition and error structure, particularly by suppressing redundant predictions. Together, these results indicate that BS-O2G realizes multi-query collaboration without the increased output redundancy that conventional O2M supervision can introduce.

\noindent\textbf{BS-O2G forms a more coherent local query family.}
We use the same affinity to construct a diagnostic top-$K$ query family for DEIM. The \emph{true-class family-query ratio} is the fraction of neighbors predicting the target class, \emph{true-class edge mass} is their total graph weight, and \emph{neighbor IoU to GT} is their affinity-weighted box overlap with the target. On this recovered subset of targets missed by DEIM but recovered by BS-O2G, all three metrics increase (Tab.~\ref{tab:query_family_quality}), indicating that the cases are accompanied by more target-consistent local query families in both semantics and geometry.

\noindent\textbf{Case study.}
Fig.~\ref{fig:paired_attention_main} presents two representative recoveries. In a scene with multiple motorcycles, DEIM predicts the correct class but produces an oversized box spanning nearby instances, whereas BS-O2G focuses on the target and localizes it more tightly. For a partially occluded apple, DEIM predicts the surrounding bowl, while BS-O2G focuses on the visible apple region and recovers the correct class and box. These cases illustrate that our method better separates nearby instances and handles occlusion.




\subsection{What Enables Effective Query Collaboration?}

\begin{table}[t]
\centering
\small
\setlength{\tabcolsep}{2.2pt}
\begin{tabularx}{\columnwidth}{@{}l
>{\centering\arraybackslash}X
>{\centering\arraybackslash}X
>{\centering\arraybackslash}X
>{\centering\arraybackslash}X|
>{\centering\arraybackslash}X@{}}
\toprule
\textbf{Variant}
& $\mathbf{Q}$
& $\mathbf{U}$
& \textbf{O2G}
& \textbf{BS}
& \textbf{AP} \\
\midrule
\multicolumn{6}{@{}c@{}}{\textbf{Training-time Ablations (24 Epochs)}} \\
\midrule
DEIM
& $\checkmark$ & -- & -- & --
& 53.0 \\
$\quad +U$
& $\checkmark$ & $\checkmark$ & -- & --
& 52.9 \\
O2G only
& $\checkmark$ & -- & $\checkmark$ & --
& 53.1 \\
\quad $+U$
& $\checkmark$ & $\checkmark$ & $\checkmark$ & --
& 53.4 \\
BS w/o O2G
& $\checkmark$ & $\checkmark$ & -- & $\checkmark$
& 52.7 \\
Random neighbors
& $\checkmark$ & $\checkmark$ & $\checkmark$ & $\checkmark$
& 52.1 \\
Eq.~\eqref{eq:graph_affinity}: first term only
& $\checkmark$ & $\checkmark$ & $\checkmark$ & $\checkmark$
& 53.2 \\
Eq.~\eqref{eq:o2g_message}: first term only
& $\checkmark$ & $\checkmark$ & $\checkmark$ & $\checkmark$
& 53.1 \\
Random BS routing
& $\checkmark$ & $\checkmark$ & $\checkmark$ & $\checkmark$
& 52.8 \\
\textbf{BS-O2G}
& $\checkmark$ & $\checkmark$ & $\checkmark$ & $\checkmark$
& \textbf{53.5} \\
\midrule
\multicolumn{6}{@{}c@{}}{\textbf{Inference-time Ablations (24 Epochs)}} \\
\midrule
\textbf{Full inference}
& $\checkmark$ & $\checkmark$ & $\checkmark$ & N/A
& \textbf{53.5} \\
w/o O2G
& $\checkmark$ & $\checkmark$ & -- & N/A
& 52.9 \\
$Q$-only ($U=0$)
& $\checkmark$ & -- & $\checkmark$ & N/A
& 53.3 \\
$U$-only ($Q=0$)
& -- & $\checkmark$ & $\checkmark$ & N/A
& 52.2 \\
\bottomrule
\end{tabularx}
\caption{Training- and inference-time ablations on COCO val2017. 
BS is training-only.}
\label{tab:training_inference_ablation}
\vspace{-0.6em}
\end{table}

\noindent\textbf{Training-time ablation.}
The upper block of Tab.~\ref{tab:training_inference_ablation} shows the staged contributions of the three components. O2G raises the baseline from 53.0 to 53.1 AP, adding the persistent basis $U$ further improves it to 53.4 AP, and BS yields the best complete result of 53.5 AP. In contrast, BS without O2G and random BS routing obtain 52.7 and 52.8 AP, respectively. These results indicate that most of the aggregate AP gain comes from O2G and $U$, while BS provides a modest, graph-dependent optimization complement within the complete framework.

\noindent\textbf{Inference-time ablation.}
The lower block of Tab.~\ref{tab:training_inference_ablation} applies ablations to the same checkpoint; BS itself is disabled at inference. Removing O2G degrades the final predictions, while the $Q$-only variant retains most of the performance and the $U$-only variant is insufficient on its own. This indicates that $Q$ carries the primary image-conditioned content, whereas $U$ serves as a complementary persistent basis.


\noindent\textbf{$U$ is shared across queries and images.}
We count the distinct queries and images routing gradients to each active basis vector. Each vector receives gradients from a median of 17 queries and 22 images, and every vector is reached by at least two queries. Thus, $U$ functions as a shared parameter bank rather than a query-specific identifier.

\begin{table}[t]
\centering
\footnotesize
\begin{tabular}{@{}lccc@{}}
\toprule
\textbf{Category pair/group} & \textbf{Learned $U$} & \textbf{Random $U$} & \textbf{Gain} \\
\midrule
Cow--sheep & .167 & .017 & +.150 \\
Related pairs & .049 & .018 & +.031 \\
Unrelated pairs & .039 & .016 & +.023 \\
\bottomrule
\end{tabular}
\caption{Total-gradient cosine under learned and random $U$ routing. Related pairs belong to the same super-category.
}
\label{tab:semantic_routing}
\vspace{-1em}
\end{table}

\noindent\textbf{$U$ carries semantically structured optimization signals.}
We first consider cow and sheep, two semantically related COCO categories. We measure whether they send gradients in similar directions to $U$ using total-gradient cosine, where a larger value means more aligned updates. As shown in Tab.~\ref{tab:semantic_routing}, 
learned $U$ routing yields a cosine of $0.167$ for cow--sheep, compared with $0.017$ under random $U$ routing. The same pattern extends across COCO: related category pairs have higher cosine than unrelated pairs under learned routing ($0.049$ vs. $0.039$) and a larger gain over random routing ($0.031$ vs. $0.023$). 
\begin{figure}[htbp]
\centering
\includegraphics[width=0.5\columnwidth]{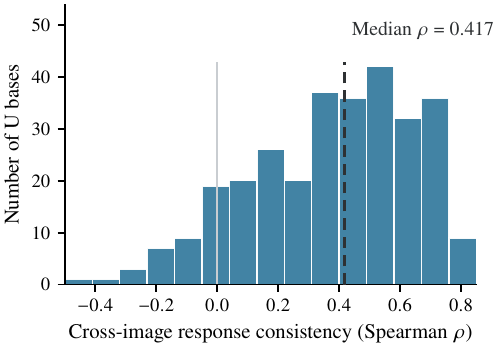}
\vspace{-0.6em}
\caption{Cross-image consistency of basis-vector class-wise logit effects. 
}
\label{fig:u_response_stability}
\vspace{-0.8em}
\end{figure}
\begin{figure}[t]
\centering
\setlength{\tabcolsep}{1pt}
\begin{tabular}{@{}cc@{}}
\makebox[0.48\columnwidth][l]{\raisebox{0.035\columnwidth}{\footnotesize (a)}\includegraphics[width=0.38\columnwidth]{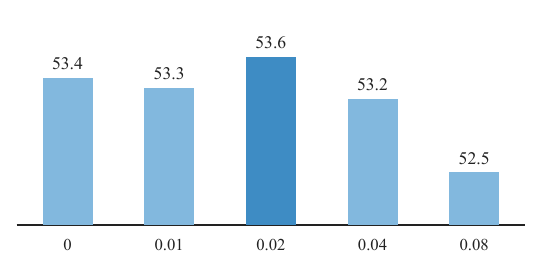}} &
\makebox[0.48\columnwidth][l]{\raisebox{0.035\columnwidth}{\footnotesize (b)}\includegraphics[width=0.38\columnwidth]{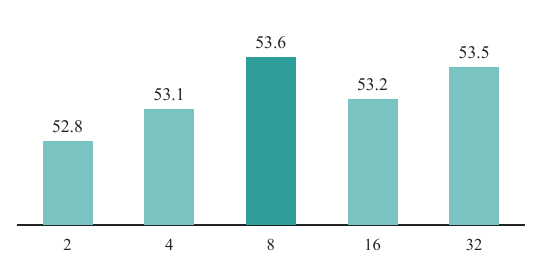}} \\
\makebox[0.48\columnwidth][l]{\raisebox{0.035\columnwidth}{\footnotesize (c)}\includegraphics[width=0.38\columnwidth]{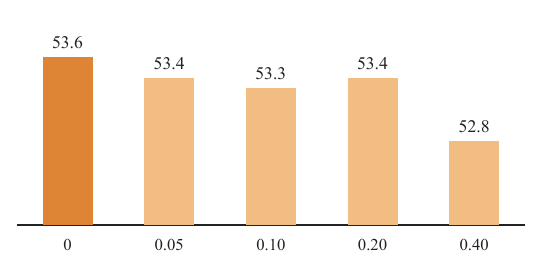}} &
\makebox[0.48\columnwidth][l]{\raisebox{0.035\columnwidth}{\footnotesize (d)}\includegraphics[width=0.38\columnwidth]{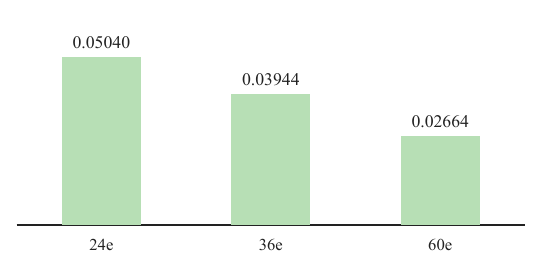}}
\end{tabular}
\vspace{-1em}
\caption{Hyperparameter analysis: (a) $\lambda_{\mathrm B}$, (b) $K$, (c) $\gamma_0$, and (d) converged $\gamma$.}
\label{fig:hyperparameter_analysis}
\vspace{-1.4em}
\end{figure}
These results show that the learned $U$ space organizes semantically related optimization signals. 

\noindent\textbf{$U$ produces repeatable class-wise logit effects across images.}
We isolate the effect of each shared basis vector by disabling O2G and evaluating the same image-conditioned query with and without that vector. For each class $c$, we quantify the class-wise logit effect as the paired change $\Delta_U(c)=z_c(Q+U)-z_c(Q)$. Across 1000 images, for example, $U_{16}$ increases the logits for handbag, dining table, and bowl by $0.077$, $0.069$, and $0.066$, respectively. These same categories are also $2.4$--$2.8\times$ overrepresented among $U_{16}$-owned detections, revealing a correspondence between category association frequency and positive logit effects. To test whether these class-wise logit effects persist across images, we compute the class-wise logit-change profile of every basis vector on two disjoint image halves. Across all basis vectors, the median split-half Spearman correlation is $0.417$ (Fig.~\ref{fig:u_response_stability}).
This consistency indicates that $U$'s class-wise logit effects include basis-vector-specific patterns that persist across images rather than arising solely from image-specific fluctuations. The $U$-swap control in Tab.~\ref{tab:u_swap_slot_control} further shows that these effects are not tied to fixed query slots. 

\noindent\textbf{Hyperparameter analysis.}
Fig.~\ref{fig:hyperparameter_analysis} examines three quantities defined above: the BS gradient-sharing weight $\lambda_{\mathrm B}$ in Eq.~\eqref{eq:bqs_gradient}, the number of graph neighbors $K$ in Eq.~\eqref{eq:graph_adjacency}, and the initialization $\gamma_0$ of the learnable O2G residual gate $\gamma$ in Eq.~\eqref{eq:o2g_update}. The best AP is obtained at $\lambda_{\mathrm B}=0.02$ and $K=8$. Initializing the gate at zero is also preferable to larger positive values, with $\gamma_0=0.4$ causing the clearest degradation. Panel (d) reports the learned gate after normal training: $\gamma$ converges to $0.05040$, $0.03944$, and $0.02664$ for the 24-, 36-, and 60-epoch schedules, respectively.

\section{Conclusion}
\label{sec:conclusion}

We identify query knowledge fragmentation in O2O DETRs, where complementary evidence is distributed across queries but only the matched owner is positively supervised. BS-O2G preserves Hungarian matching and the O2O objective while coordinating O2G forward calibration and BS backward sharing over a shared prediction-aware graph. Across COCO and CrowdHuman, BS-O2G consistently improves diverse detectors with faster convergence, negligible parameter/FLOP growth,  and no extra positive assignments. 


\bibliography{aaai2027,detr}
\clearpage
\appendix
\setcounter{secnumdepth}{2}
\setcounter{page}{1}
\renewcommand{\thepage}{S\arabic{page}}
\noindent\textbf{Appendix Overview}\par\smallskip
\begingroup
\small
\setlength{\tabcolsep}{0pt}
\noindent\begin{tabularx}{\columnwidth}{@{}Xr@{}}
\textbf{\ref{app:extended_method_details}\enspace Extended Method Details} & p.~\pageref{app:extended_method_details} \\
\quad \ref{app:bqs_derivation}\enspace BS Gradient Transformation & p.~\pageref{app:bqs_derivation} \\
\textbf{\ref{app:theoretical_proofs}\enspace Proofs of the Theoretical Results} & p.~\pageref{app:theoretical_proofs} \\
\quad \ref{app:proof_bqs_descent}\enspace Local Descent under Directed BS Routing & p.~\pageref{app:proof_bqs_descent} \\
\textbf{\ref{app:reproducibility}\enspace Reproducibility and Additional Measurements} & p.~\pageref{app:reproducibility} \\
\quad \ref{app:fragmentation_setup}\enspace Figure 1 Diagnostic Protocol & p.~\pageref{app:fragmentation_setup} \\
\quad \ref{app:crowdhuman_pr}\enspace CrowdHuman Scale-Wise PR Profiles & p.~\pageref{app:crowdhuman_pr} \\
\textbf{\ref{app:mechanism_analysis}\enspace Additional Mechanism Analysis} & p.~\pageref{app:mechanism_analysis} \\
\quad \ref{app:role_bs}\enspace Role of BS & p.~\pageref{app:role_bs} \\
\quad \ref{app:basis_response_analysis}\enspace Persistent Query-Basis Response Analysis & p.~\pageref{app:basis_response_analysis} \\
\quad \ref{app:neighbor_intervention}\enspace Neighbor Intervention & p.~\pageref{app:neighbor_intervention} \\
\textbf{\ref{app:qualitative_results}\enspace Additional Qualitative Results} & p.~\pageref{app:qualitative_results} \\
\end{tabularx}
\endgroup

\paragraph{AI Use Disclosure.}
Generative AI tools were used to assist with language editing, code
debugging, and figure drafting. The authors reviewed and verified all
AI-assisted outputs and take full responsibility for the content of
this manuscript.

\section{Extended Method Details}
\label{app:extended_method_details}

\subsection{Complete BS-O2G Training and Inference Procedure}
\label{app:bqs_o2g_algorithm}

Algorithm~\ref{alg:bqs_o2g_complete} gives the complete procedure. All graph operations act only on normal queries; denoising queries, when present, follow the original detector path. The same detached adjacency is used by O2G in the forward pass and by BS in the backward pass. Consequently, training preserves the original Hungarian matching and detection objective, while inference retains the learned basis $U$ and O2G calibration but performs no BS gradient routing.

\begin{algorithm}[htbp]
\caption{Complete BS-O2G training and inference}
\label{alg:bqs_o2g_complete}
\begin{algorithmic}[1]
\REQUIRE Image batch $X$; targets $Y$ if training; persistent basis $U$; neighborhood size $K$; temperature $\tau$; O2G gate $\gamma$; BS coefficient $\lambda_{\mathrm B}$
\ENSURE Calibrated predictions $(\widehat B,\widehat Z)$; updated parameters if training
\STATE Obtain image-conditioned normal queries $Q(X)$ and reference boxes $R(X)$ from encoder proposals
\STATE Form normal-query inputs $Q^0(X)\leftarrow Q(X)+U$; leave denoising-query inputs unchanged
\STATE Decode $Q^0(X)$ to obtain final normal-query features $H$ and preliminary predictions $(B,Z)$
\STATE Compute class probabilities $P\leftarrow\sigma(Z)$ and affinities $s_{ij}$ using Eq.~\eqref{eq:graph_affinity}
\FOR{each normal query $i$}
    \STATE $\mathcal N_i\leftarrow\operatorname{TopK}_{j\ne i}(s_{ij},K)$
    \STATE $A_{ij}\leftarrow\operatorname{softmax}_{j\in\mathcal N_i}(s_{ij}/\tau)$ and detach $A_{i:}$
    \FOR{each $j\in\mathcal N_i$}
        \STATE Compute relative message $v_{ij}$ using Eq.~\eqref{eq:o2g_message}
    \ENDFOR
    \STATE $\widetilde h_i\leftarrow h_i+\gamma W_o\sum_{j\in\mathcal N_i}A_{ij}v_{ij}$
\ENDFOR
\STATE Recompute $(\widehat B,\widehat Z)$ from $\widetilde H$ using the original prediction heads
\IF{training}
    \STATE Set $\lambda_{\mathrm B}$ by the delayed warm-up schedule
    \STATE Register the basis-gradient hook $G_U\mapsto(1-\lambda_{\mathrm B})G_U+\lambda_{\mathrm B}A^\top G_U$
    \STATE Compute the original O2O Hungarian assignment and detector loss $\mathcal L(\widehat B,\widehat Z;Y)$
    \STATE Backpropagate $\mathcal L$ and update all parameters with the transformed basis gradient
\ELSE
    \STATE Skip the BS hook and return the O2G-calibrated predictions
\ENDIF
\STATE \textbf{return} $(\widehat B,\widehat Z)$
\end{algorithmic}
\end{algorithm}

The hook in Algorithm~\ref{alg:bqs_o2g_complete} is attached only after the current forward pass has constructed $A$, avoiding any look-ahead graph dependency in the decoder input. The clean-form transformation shown above is Eq.~\eqref{eq:bqs_gradient}; the numerical sanitizers and hook-level implementation are detailed in the hook-based implementation subsection below.

\subsection{Relation to Decoder Self-Attention}
\label{app:self_attention_graph}

Decoder self-attention already enables queries to interact, but interaction weights are not equivalent to the detection-aware relation required by BS-O2G. The distinction is one of semantics and operational role rather than expressive capacity: self-attention describes layer-specific latent information flow, whereas our graph must measure relations between decoded detection candidates and provide one sparse, detached operator that can be reused for both forward calibration and backward sharing.

\paragraph{Self-attention measures latent read dependency.}
For decoder layer $\ell$ and attention head $r$, let $x_i^{(\ell)}$ and $\pi_i^{(\ell)}$ denote the content and positional states of query $i$. Vanilla self-attention computes
\begin{align}
q_i^{(\ell,r)}&=W_Q^r(x_i^{(\ell)}+\pi_i^{(\ell)}),\\
k_j^{(\ell,r)}&=W_K^r(x_j^{(\ell)}+\pi_j^{(\ell)}),\\
\alpha_{ij}^{(\ell,r)}
&=\operatorname{softmax}_{j}\!\left(
\frac{(q_i^{(\ell,r)})^\top k_j^{(\ell,r)}}{\sqrt{d_r}}
\right),\\
y_i^{(\ell,r)}
&=\sum_j\alpha_{ij}^{(\ell,r)}W_V^r x_j^{(\ell)}.
\label{eq:app_self_attention}
\end{align}
Thus, $\alpha_{ij}^{(\ell,r)}$ quantifies how strongly query $i$ reads the value of query $j$ for a particular head and layer. It is a dense, row-normalized routing coefficient learned only through the downstream objective. A large coefficient can support cooperative evidence aggregation, competitive duplicate suppression, or another latent computation; its magnitude alone does not identify which interpretation applies. In particular, vanilla self-attention does not explicitly require $\alpha_{ij}^{(\ell,r)}$ to reflect box overlap, class agreement, or whether two candidates contain complementary evidence for the same object.

\paragraph{The attention weights are not sufficient for the final detection relation.}
In the decoder used by our baseline, self-attention precedes cross-attention and the feed-forward network, after which the prediction heads produce the decoded boxes and class distributions. Abstractly, the self-attention matrix and final prediction state can therefore be written as
\begin{align}
A_{\mathrm{SA}}^{(\ell)}
&=f_{\mathrm{SA}}(X^{(\ell)},\Pi^{(\ell)}),\\
(H^{(\ell)},B^{(\ell)},P^{(\ell)})
&=g(X^{(\ell)},\Pi^{(\ell)},R^{(\ell)},M).
\label{eq:app_attention_stage_mismatch}
\end{align}
where $R^{(\ell)}$ denotes the reference state and $M$ the image memory used by cross-attention. Holding $X^{(\ell)}$, $\Pi^{(\ell)}$, and $R^{(\ell)}$ fixed while changing $M$ leaves $A_{\mathrm{SA}}^{(\ell)}$ unchanged, yet can change the cross-attention output and hence the final features, boxes, and class distributions. Consequently, no function of the self-attention weights alone can, in general, recover a relation defined on the final detection state. This is an architectural insufficiency statement, not a claim that self-attention learns no useful query interaction.

\paragraph{The prediction-aware graph makes the required semantics explicit.}
Equation~\eqref{eq:graph_affinity} constructs each edge after decoding from final query-feature similarity, predicted-box IoU, and class-distribution similarity. Its row-wise top-$K$ sparsification in Eq.~\eqref{eq:graph_adjacency} selects a fixed-size local neighborhood and removes self-edges. The resulting $A_{ij}$ therefore has an explicit interpretation under the current prediction state: it measures how suitable query $j$ is as a detection-aware neighbor of query $i$. Moreover, vanilla self-attention sends the receiver-independent value $W_V^r x_j^{(\ell)}$ from query $j$, modulated by a scalar coefficient. By contrast, Eq.~\eqref{eq:o2g_message} constructs the edge-conditioned vector
\begin{equation}
v_{ij}=\phi_m([h_j-h_i;\rho(b_i,b_j);\operatorname{IoU}(b_i,b_j);p_j-p_i]),
\label{eq:app_edge_conditioned_message}
\end{equation}
so the message sent by the same query $j$ changes with the receiving query $i$ through relative feature, geometry, overlap, and class evidence. O2G thus separates \emph{which neighbor to use}, encoded by $A_{ij}$, from \emph{what relational evidence to send}, encoded by $v_{ij}$.

\paragraph{A self-attention map is not a reusable backward-routing operator.}
Even when one self-attention matrix is extracted, its ordinary backward pass is not the transposed routing used by BS. For $Y=A_{\mathrm{SA}}(H)V(H)$ and an arriving gradient $G=\partial\mathcal L/\partial Y$, the differential is
\begin{equation}
\mathrm dY=(\mathrm dA_{\mathrm{SA}})V+A_{\mathrm{SA}}\,\mathrm dV.
\label{eq:app_attention_differential}
\end{equation}
Hence, the gradient with respect to $H$ contains both a value-path term induced by $A_{\mathrm{SA}}^\top G$ and additional score-path terms induced by the dependence of $A_{\mathrm{SA}}$ on the query and key projections. It is therefore not equal to a clean $A_{\mathrm{SA}}^\top G$ transformation. Self-attention also yields a different dense matrix for every layer and head, with no canonical choice to serve as a shared relation operator. BS instead detaches one post-decoder adjacency and explicitly applies Eq.~\eqref{eq:bqs_gradient}, yielding the controlled transformation $(1-\lambda_{\mathrm B})G_U+\lambda_{\mathrm B}A^\top G_U$ on persistent query basis vectors. The same directed edges consequently have complementary forward and backward meanings: if query $i$ reads detection-aware evidence from query $j$ in O2G, query $i$'s gradient can be routed back to basis vector $u_j$ in BS.

\paragraph{Scope of the distinction.}
Self-attention could be modified with prediction-derived biases, sparse neighborhoods, relative edge features, and a detached adjacency reused outside the attention layer. Such a construction may approximate the functionality above, but it is no longer vanilla decoder self-attention; it effectively becomes a prediction-aware graph-attention operator. BS-O2G therefore does not replace the decoder's existing query interaction. It complements it with an output-aligned graph whose edge semantics, sparsity, relative messages, and optimization routing are explicitly defined for detection.

\subsection{Derivation of the BS Gradient Transformation}
\label{app:bqs_derivation}

We provide the detailed derivation of Eq.~\eqref{eq:bqs_gradient}. We consider one image and omit the image index of $A$ for clarity. Let $U\in\mathbb{R}^{N\times d}$ be the expanded query basis, $A\in\mathbb{R}^{N\times N}$ the detached adjacency matrix, and $\lambda_{\mathrm B}$ the BS sharing coefficient. During this derivation, both $A$ and $\lambda_{\mathrm B}$ are treated as constants.

\paragraph{Straight-through forward identity.}
The graph-composed residual and the straight-through basis are
\begin{equation}
\Delta U=AU-U=(A-I_N)U,
\label{eq:appendix_bqs_residual}
\end{equation}
\begin{equation}
U^{\mathrm{BS}}
=U+\lambda_{\mathrm B}\Delta U
-\operatorname{sg}\!\left(\lambda_{\mathrm B}\Delta U\right),
\label{eq:appendix_bqs_st}
\end{equation}
where $I_N$ is the $N\times N$ identity matrix. The stop-gradient operator preserves its input value in the forward pass but has zero derivative. Therefore,
\begin{equation}
U^{\mathrm{BS}}
\overset{\mathrm{forward}}{=}
U+\lambda_{\mathrm B}\Delta U-\lambda_{\mathrm B}\Delta U
=U.
\label{eq:appendix_bqs_forward}
\end{equation}
Thus, BS leaves the numerical decoder input unchanged while retaining an additional graph-dependent derivative path.

\paragraph{Backward Jacobian.}
Because $A$ is detached, its differential is zero. Differentiating Eq.~\eqref{eq:appendix_bqs_residual} gives
\begin{equation}
\mathrm{d}(\Delta U)
=A\,\mathrm{d}U-\mathrm{d}U
=(A-I_N)\mathrm{d}U.
\label{eq:appendix_bqs_residual_diff}
\end{equation}
The stop-gradient term in Eq.~\eqref{eq:appendix_bqs_st} contributes no derivative, and hence
\begin{align}
\mathrm{d}U^{\mathrm{BS}}
&=\mathrm{d}U+\lambda_{\mathrm B}\,\mathrm{d}(\Delta U)\nonumber\\
&=\left[I_N+\lambda_{\mathrm B}(A-I_N)\right]\mathrm{d}U\nonumber\\
&=\left[(1-\lambda_{\mathrm B})I_N+\lambda_{\mathrm B}A\right]\mathrm{d}U.
\label{eq:appendix_bqs_jacobian}
\end{align}

Let $G_U=\partial\mathcal{L}/\partial U^{\mathrm{BS}}$ denote the gradient arriving at the straight-through output. Using the Frobenius inner product $\langle X,Y\rangle_F=\operatorname{tr}(X^\top Y)$, the loss differential is
\begin{align}
\mathrm{d}\mathcal{L}
&=\left\langle G_U,\mathrm{d}U^{\mathrm{BS}}\right\rangle_F\nonumber\\
&=\left\langle G_U,
\left[(1-\lambda_{\mathrm B})I_N+\lambda_{\mathrm B}A\right]\mathrm{d}U
\right\rangle_F\nonumber\\
&=\left\langle
\left[(1-\lambda_{\mathrm B})I_N+\lambda_{\mathrm B}A\right]^\top G_U,
\mathrm{d}U
\right\rangle_F.
\label{eq:appendix_bqs_loss_diff}
\end{align}
Therefore, the gradient with respect to the learnable basis is
\begin{align}
G_U^{\mathrm{BS}}
&=\frac{\partial\mathcal{L}}{\partial U}\nonumber\\
&=\left[(1-\lambda_{\mathrm B})I_N+\lambda_{\mathrm B}A\right]^\top G_U\nonumber\\
&=(1-\lambda_{\mathrm B})G_U+\lambda_{\mathrm B}A^\top G_U,
\label{eq:appendix_bqs_gradient}
\end{align}
which is Eq.~\eqref{eq:bqs_gradient}.

\paragraph{Query-wise interpretation.}
Let $g_i$ denote the $i$-th row of $G_U$. Expanding Eq.~\eqref{eq:appendix_bqs_gradient} for basis vector $u_j$ gives
\begin{equation}
g_j^{\mathrm{BS}}
=(1-\lambda_{\mathrm B})g_j
+\lambda_{\mathrm B}\sum_{i=1}^{N}A_{ij}g_i.
\label{eq:appendix_bqs_querywise}
\end{equation}
Hence, if query $i$ reads query $j$ with forward weight $A_{ij}$, the gradient of query $i$ contributes to basis vector $u_j$ with weight $\lambda_{\mathrm B}A_{ij}$. The transpose in Eq.~\eqref{eq:bqs_gradient} therefore follows directly from reversing the dependency direction during backpropagation. This transformation changes only the gradient assigned to the normal-query basis vectors; it does not introduce a new loss or modify the Hungarian assignment.

\subsubsection{Hook-Based Implementation without Look-Ahead}
\label{app:bqs_hook}

The straight-through expression in Eq.~\eqref{eq:appendix_bqs_st} characterizes the BS Jacobian but is not evaluated as a graph-composed basis before decoding. At that point, $A$ is not yet available because it is constructed from the decoded features and preliminary predictions. Directly using $AU$ in the decoder input would therefore introduce a look-ahead dependency between the query basis and a graph produced later in the same forward pass.

We avoid this dependency with a backward hook. During the forward pass, the expanded basis $U_{\mathrm{exp}}$ is added to $Q(x)$ without graph composition and retained in the computation graph. After the decoder produces $H$, $B$, and $Z$, we construct and detach $A$, and then register the following hook on $U_{\mathrm{exp}}$ before backpropagation:
\begin{equation}
\mathcal{H}_{A}(G)
=(1-\lambda_{\mathrm B})G+\lambda_{\mathrm B}A^\top G.
\label{eq:appendix_bqs_hook}
\end{equation}
For numerical safety, the code first applies the elementwise sanitizer $S_G(G)=\operatorname{clip}(\operatorname{nan\_to\_num}(G),[-10^4,10^4])$ and returns $\mathcal H_A(S_G(G))$. Equation~\eqref{eq:appendix_bqs_hook} is exact for the raw gradient whenever this sanitizer is inactive; the basis path has an analogous sanitizer whose Jacobian is the identity in the clean region. When autograd reaches $U_{\mathrm{exp}}$, the hook replaces the arriving gradient with this routed value. The resulting per-image gradients are then accumulated into the shared parameter $U$ by standard backpropagation, after which the optimizer performs the parameter update. Algorithm~\ref{alg:bqs_hook} summarizes this procedure.

\begin{algorithm}[t]
\caption{Hook-based BS gradient routing}
\label{alg:bqs_hook}
\begin{algorithmic}[1]
\REQUIRE Image-conditioned queries $Q(x)$, query basis $U$, sharing coefficient $\lambda_{\mathrm B}$
\STATE Expand $U$ over the batch to obtain $U_{\mathrm{exp}}$
\STATE Form the decoder input $Q^0(x)\leftarrow Q(x)+U_{\mathrm{exp}}$
\STATE Decode $Q^0(x)$ to obtain $H$ and preliminary predictions $(B,Z)$
\STATE Construct $A\leftarrow\operatorname{sg}(\operatorname{Graph}(H,B,Z))$
\STATE Apply O2G to obtain final predictions $(\widehat B,\widehat Z)$
\STATE Register $\mathcal{H}_{A}$ from Eq.~\eqref{eq:appendix_bqs_hook} on $U_{\mathrm{exp}}$
\STATE Compute the original detection loss $\mathcal{L}(\widehat B,\widehat Z;Y)$
\STATE Backpropagate $\mathcal{L}$; the hook transforms the gradient arriving at $U_{\mathrm{exp}}$
\STATE Update $U$ with the accumulated transformed gradient
\end{algorithmic}
\end{algorithm}

The hook is registered only for normal-query basis vectors; denoising queries bypass both the shared graph and BS. This implementation uses the graph produced by the current forward pass without requiring an additional forward pass, premature parameter update, or gradient through the graph construction itself.

\section{Proofs of the Theoretical Results}
\label{app:theoretical_proofs}

All statements below use exact arithmetic. The descent result concerns the clean region in which the implementation's gradient and basis sanitizers act as the identity.

\subsection{Proof of Theorem~\ref{thm:o2g_mse}}
\label{app:proof_o2g_mse}

Fix a target query $i$ and let $h_i^\star$ denote its latent object representation. We model $h_i=h_i^\star+\varepsilon_i$ and $h_j=h_i^\star+\delta_{ij}+\varepsilon_j$ for $j\in\mathcal N_i$. Conditional on $A_{i:}$, the noise vectors are independent, zero mean, and have covariance $\nu^2I_d$, while $\delta_{ij}$ is deterministic. Recall that $\bar\delta_i=\sum_{j\in\mathcal N_i}A_{ij}\delta_{ij}$.

Since $\sum_{j\in\mathcal N_i}A_{ij}=1$, Eq.~\eqref{eq:o2g_linear_slice} can be written as
\begin{equation}
\widetilde h_i^{\mathrm{lin}}
=(1-\gamma)h_i+\gamma\sum_{j\in\mathcal N_i}A_{ij}h_j.
\label{eq:app_o2g_convex_form}
\end{equation}
Substituting $h_i=h_i^\star+\varepsilon_i$ and $h_j=h_i^\star+\delta_{ij}+\varepsilon_j$ gives
\begin{equation}
\widetilde h_i^{\mathrm{lin}}-h_i^\star
=(1-\gamma)\varepsilon_i
+\gamma\bar\delta_i
+\gamma\sum_{j\in\mathcal N_i}A_{ij}\varepsilon_j.
\label{eq:app_o2g_error_decomp}
\end{equation}
Conditional on $A_{i:}$, the noise terms are independent and zero mean, so every cross term in the squared norm has zero conditional expectation. Moreover, $\mathbb E[\|\varepsilon_j\|_2^2\mid A_{i:}]=d\nu^2$. Let $\mathcal E_i(\gamma)$ denote the conditional MSE on the left-hand side of Eq.~\eqref{eq:o2g_mse}. Hence
\begin{align}
\mathcal E_i(\gamma)
&=(1-\gamma)^2d\nu^2
+\gamma^2\|\bar\delta_i\|_2^2
+\gamma^2\!\sum_{j\in\mathcal N_i}A_{ij}^2d\nu^2\nonumber\\
&=(1-\gamma)^2d\nu^2
+\frac{\gamma^2}{K_i^{\mathrm{eff}}}d\nu^2
+\gamma^2\|\bar\delta_i\|_2^2,
\label{eq:o2g_mse}
\end{align}
which proves Eq.~\eqref{eq:o2g_mse}. Subtracting the baseline MSE $d\nu^2$ yields
\begin{equation}
\Delta\mathcal E_i(\gamma)
=-2\gamma d\nu^2
+\gamma^2\!\left[d\nu^2(1+1/K_i^{\mathrm{eff}})+\|\bar\delta_i\|_2^2\right].
\label{eq:app_o2g_mse_difference}
\end{equation}
Let $M_i=d\nu^2(1+1/K_i^{\mathrm{eff}})+\|\bar\delta_i\|_2^2>0$. Then $\Delta\mathcal E_i(\gamma)=\gamma(-2d\nu^2+\gamma M_i)$, which is negative exactly on the interval in Eq.~\eqref{eq:o2g_improvement_interval}. Differentiating gives the unique minimizer $\gamma^*=d\nu^2/M_i$. If $\delta_{ij}=0$ and $A_{ij}=1/K$ for $j\in\mathcal N_i$, then $K_i^{\mathrm{eff}}=K$ and the minimum MSE divided by $d\nu^2$ is $1/(K+1)$. This completes the proof.

\subsection{Proof of Theorem~\ref{thm:bqs_contraction}}
\label{app:proof_bqs_contraction}

Since $A$ is symmetric, $A^\top=A$, and the spectral theorem gives an orthonormal eigenbasis $\{v_r\}_{r=1}^N$. Row stochasticity and connectedness imply $v_1=N^{-1/2}\mathbf{1}_N$, $\mu_1=1$, and $\mu_2<1$. The routing operator $P_{\mathrm B}=(1-\lambda_{\mathrm B})I_N+\lambda_{\mathrm B}A^\top$ therefore has the same eigenvectors, with eigenvalues
\begin{equation}
p_r=1-\lambda_{\mathrm B}(1-\mu_r).
\label{eq:app_bqs_filter_eigenvalue}
\end{equation}
For the consensus mode, $p_1=1$, which proves mean preservation. For $r\ge2$, the conditions $\mu_r\ge-1$ and $0<\lambda_{\mathrm B}\le1/2$ give
\begin{equation}
0\le p_r\le1-\lambda_{\mathrm B}(1-\mu_2)=\rho_{\mathrm B}<1.
\label{eq:app_bqs_filter_bound}
\end{equation}
Because $G_{U,\perp}$ has zero mean, it has the expansion $G_{U,\perp}=\sum_{r=2}^Nv_rc_r^\top$ for vectors $c_r\in\mathbb R^d$. Orthogonality then yields
\begin{align}
\|P_{\mathrm B}G_{U,\perp}\|_F^2
&=\sum_{r=2}^Np_r^2\|c_r\|_2^2\nonumber\\
&\le\rho_{\mathrm B}^2\sum_{r=2}^N\|c_r\|_2^2
=\rho_{\mathrm B}^2\|G_{U,\perp}\|_F^2.
\label{eq:app_bqs_contraction_proof}
\end{align}
Taking square roots proves Eq.~\eqref{eq:bqs_contraction} and completes the proof.

\subsection{Additional Result: Local Descent Preservation under Directed BS Routing}
\label{app:proof_bqs_descent}

For a fixed image $x$, let $\ell_x(U)$ denote the detection loss with the detached graph frozen, $G_U=\nabla_U\ell_x(U)\neq0$, and $G_U^{\mathrm{BS}}=[(1-\lambda_{\mathrm B})I_N+\lambda_{\mathrm B}A^\top]G_U$. We consider the clean update region in which the numerical gradient and basis sanitizers are inactive.

\begin{theorem}[Local descent under directed BS routing]
\label{thm:bqs_descent}
For any nonnegative row-stochastic $A$, if
\begin{equation}
0\le\lambda_{\mathrm B}<\frac{1}{1+\sqrt N},
\label{eq:bqs_safe_lambda}
\end{equation}
then
\begin{equation}
\begin{aligned}
\langle G_U,G_U^{\mathrm{BS}}\rangle_F
&\ge c_{\mathrm{align}}\|G_U\|_F^2,\\
c_{\mathrm{align}}
&=1-\lambda_{\mathrm B}(1+\sqrt N)>0.
\end{aligned}
\label{eq:bqs_descent_alignment}
\end{equation}
Moreover, if $\ell_x$ is locally $L$-smooth and $c_{\mathrm{route}}=1-\lambda_{\mathrm B}+\lambda_{\mathrm B}\sqrt N$, every $0<\eta<2c_{\mathrm{align}}/(Lc_{\mathrm{route}}^2)$ satisfies $\ell_x(U-\eta G_U^{\mathrm{BS}})<\ell_x(U)$.
\end{theorem}

For the default setting of $N=300$ normal queries and target routing coefficient $\lambda_{\mathrm B}=0.02$, the conservative alignment margin is $c_{\mathrm{align}}\approx0.6336$. This quantity is an unnormalized worst-case inner-product lower bound, rather than a cosine similarity or a retained-gradient ratio. The result establishes that, for a fixed image and frozen detached graph in the clean update region, the raw BS-routed gradient remains a strict descent direction under a sufficiently small gradient step. It does not guarantee descent after mini-batch gradient accumulation, characterize the AdamW update, imply global convergence, or establish an improvement in detection AP. Theorem~\ref{thm:bqs_contraction} provides a graph low-pass interpretation under a balanced symmetric limit, whereas Theorem~\ref{thm:bqs_descent} establishes local descent preservation for the directed row-stochastic routing operator used in practice.

\paragraph{Proof.}
Because $A$ is nonnegative and row-stochastic, $\|A\|_\infty=1$. Its maximum column sum is at most $N$, so $\|A\|_1\le N$. The induced-norm inequality therefore gives
\begin{equation}
\|A\|_2\le\sqrt{\|A\|_1\|A\|_\infty}\le\sqrt N.
\label{eq:app_bqs_spectral_bound}
\end{equation}
Using Cauchy--Schwarz and Eq.~\eqref{eq:app_bqs_spectral_bound},
\begin{align}
\langle G_U,G_U^{\mathrm{BS}}\rangle_F
&=(1-\lambda_{\mathrm B})\|G_U\|_F^2
+\lambda_{\mathrm B}\langle G_U,A^\top G_U\rangle_F\nonumber\\
&\ge(1-\lambda_{\mathrm B})\|G_U\|_F^2\nonumber\\
&\quad-\lambda_{\mathrm B}\|G_U\|_F\|A^\top G_U\|_F\nonumber\\
&\ge[1-\lambda_{\mathrm B}(1+\sqrt N)]\|G_U\|_F^2
\nonumber\\
&=c_{\mathrm{align}}\|G_U\|_F^2.
\label{eq:app_bqs_alignment_proof}
\end{align}
Equation~\eqref{eq:bqs_safe_lambda} makes $c_{\mathrm{align}}$ strictly positive, and therefore $-G_U^{\mathrm{BS}}$ is a strict descent direction. In addition,
\begin{equation}
\|G_U^{\mathrm{BS}}\|_F
\le(1-\lambda_{\mathrm B}+\lambda_{\mathrm B}\sqrt N)\|G_U\|_F
=c_{\mathrm{route}}\|G_U\|_F.
\label{eq:app_bqs_direction_norm}
\end{equation}
By $L$-smoothness,
\begin{align}
&\ell_x(U-\eta G_U^{\mathrm{BS}})-\ell_x(U)\nonumber\\
&\quad\le-\eta\langle G_U,G_U^{\mathrm{BS}}\rangle_F
+\frac{L\eta^2}{2}\|G_U^{\mathrm{BS}}\|_F^2\nonumber\\
&\quad\le-\eta c_{\mathrm{align}}\|G_U\|_F^2
+\frac{L\eta^2c_{\mathrm{route}}^2}{2}\|G_U\|_F^2.
\label{eq:app_bqs_descent_lemma}
\end{align}
The right-hand side is strictly smaller than $\ell_x(U)$ whenever $0<\eta<2c_{\mathrm{align}}/(Lc_{\mathrm{route}}^2)$, proving the theorem.

\section{Reproducibility and Additional Measurements}
\label{app:reproducibility}

\subsection{Datasets and Availability}
\label{app:datasets}

We use two publicly available benchmarks with complementary roles. MS-COCO~\cite{DBLP:conf/eccv/LinMBHPRDZ14} evaluates general multi-class object detection across diverse scenes and object scales, while CrowdHuman~\cite{DBLP:journals/corr/abs-1805-00123} provides a denser and more occlusion-heavy pedestrian setting for testing whether BS-O2G transfers beyond COCO and remains effective under stronger query competition.

For MS-COCO, we train on \texttt{train2017} with 118,287 images and evaluate on \texttt{val2017} with 5,000 images over 80 object categories. We follow the standard COCO bounding-box protocol and report AP and AR across IoU thresholds and object scales. For CrowdHuman, we use the official training split with 15,000 images and validation split with 4,370 images. Following the DEIM protocol, we convert the annotations to COCO format and train a one-class person detector using full-body bounding boxes.

Both datasets are available through their official public distributions under their respective terms of use. We introduce no new dataset and use no private or otherwise unavailable data.

\subsection{Evaluation Metrics}
\label{app:evaluation_metrics}

For both MS-COCO and the COCO-formatted CrowdHuman full-body annotations, we use the standard COCO bounding-box evaluator. AP averages precision over IoU thresholds from $0.50$ to $0.95$ in increments of $0.05$, while AP$_{50}$ and AP$_{75}$ separately expose performance under looser and stricter localization criteria. AP$_S$, AP$_M$, and AP$_L$ report the same measure for small, medium, and large objects. We use AP as the primary detection metric because it jointly reflects classification and localization quality across operating points, and use its IoU- and scale-specific variants to determine whether an overall change is concentrated at a particular localization strictness or object scale.

We additionally report AR$_{10}$ and AR$_{100}$, the maximum recall obtained with at most 10 and 100 detections per image, together with scale-specific AR. These metrics complement AP by measuring whether BS-O2G recovers objects that would otherwise be missed, which is directly relevant to query knowledge fragmentation. The custom diagnostics used for mechanism analysis are defined at their first use: Fig.~\ref{fig:knowledge_fragmentation} uses winner agreement, evidence dispersion, and owner coverage as formalized in Eq.~\eqref{eq:appendix_fragmentation_metrics}; the set-level analysis distinguishes duplicate and localization false positives; and the query-family and routing analyses use explicitly defined class-consistency, edge-mass, IoU, Jaccard, and gradient-cosine measures. Together, these diagnostics test coverage, redundancy, neighborhood coherence, and optimization sharing rather than treating aggregate AP as mechanism evidence by itself.

For computational cost, we report parameter count, FLOPs at $640\times640$ resolution, batch-1 latency, throughput, and CUDA peak allocated memory. Latency and throughput measure practical inference cost, while the forward-and-backward time and peak training memory isolate training overhead; their warm-up, synchronization, repetition, and aggregation protocols are specified in Appendix~\ref{app:additional_efficiency}.

\subsection{Randomness Control}
\label{app:randomness_control}

We control training randomness through an explicit base seed passed by the command-line argument \texttt{--seed}. The main run uses seed 0, and the additional replicated runs use seeds 42 and 1027. In distributed training, a process with global rank $r$ uses the effective seed $s+r$, where $s$ is the selected base seed. This effective seed initializes Python's \texttt{random} module, NumPy, the PyTorch CPU random-number generator, and all CUDA random-number generators through \texttt{random.seed}, \texttt{numpy.random.seed}, \texttt{torch.manual\_seed}, and \texttt{torch.cuda.manual\_seed\_all}, respectively. PyTorch's worker-seeding mechanism then assigns reproducible worker-specific seeds derived from the seeded training process.

At every epoch, we call \texttt{set\_epoch(epoch)} for the training data loader and, under distributed training, for its \texttt{DistributedSampler}. This makes epoch-dependent data shuffling and the dataset/collation augmentation schedule reproducible for a fixed seed. To replicate a run, we keep the seed, dataset split, model and training configuration, number of GPUs, and software environment unchanged. We do not explicitly enforce CUDA deterministic algorithms; consequently, fixed-seed runs are reproducible under the same environment, but exact bitwise identity is not guaranteed across different hardware or software versions.

\subsection{Computing Environment}
\label{app:computing_environment}

All reproduced training runs were conducted on a server with two Intel Xeon Platinum 8558 processors (48 cores per socket, 96 physical cores and 192 hardware threads in total; 80 logical CPUs were visible to the jobs) and eight NVIDIA H200 GPUs. Each GPU provides 143,771 MiB (approximately 140.4 GiB; nominally 141 GB) of memory. The operating system was CentOS Linux 7 (Core).

The software environment used Python 3.10.12, PyTorch 2.7.0+cu128, and TorchVision 0.22.0+cu128. PyTorch was compiled against CUDA 12.8, and the installed CUDA Toolkit/\texttt{nvcc} version was 12.8 (V12.8.93). The NVIDIA driver version was 535.216.03; \texttt{nvidia-smi} reported CUDA compatibility version 12.2. We report the build, toolkit, driver, and compatibility versions separately to avoid conflating these distinct CUDA version indicators.

\subsection{Additional Implementation Details}
\label{app:implementation_details}

The default DEIM-RT-DETRv2-R50 model contains six decoder layers, a hidden dimension of 256, 300 normal queries, and 100 denoising queries. We use a flat-cosine learning-rate schedule with 2,000 warm-up iterations, gradient clipping with a maximum norm of $0.1$, exponential moving average, and automatic mixed precision. Following DEIM, training augmentation includes Mosaic, MixUp, random photometric distortion, random zoom-out, random IoU crop, horizontal flipping, and multi-scale resizing; strong augmentations are disabled during the final two epochs.

For cross-architecture evaluation, we integrate BS-O2G into DINO, RT-DETRv2, and D-FINE with ResNet-50, ResNet-101, HGNetV2-B4, and HGNetV2-B5 backbones while retaining their detector-specific configurations. The feature, box, and class terms use unit external coefficients in the graph affinity, with the feature term retaining the $1/\sqrt d$ scaling in Eq.~\eqref{eq:graph_affinity}; no graph dropout is used. O2G and BS operate only on normal object queries; denoising queries are excluded from graph construction and gradient routing. BS-O2G introduces neither additional positive assignments nor auxiliary detection losses.

\subsection{Figure 1 Diagnostic Protocol}
\label{app:fragmentation_setup}

Fig.~\ref{fig:knowledge_fragmentation} reports a post-hoc, ground-truth-aware analysis of frozen baseline detectors, rather than an additional training objective. For COCO, we evaluate the EMA checkpoint of a 60-epoch DEIM-RT-DETRv2-R50 baseline on train2017 using deterministic validation transforms. Based on post-resize box areas, the diagnostic contains 213,127 small, 299,629 medium, and 337,191 large valid non-crowd instances. The corresponding CrowdHuman setting is specified below.

For each ground-truth object $g$, we use the predictions from the final decoder layer and recompute the detector's original Hungarian cost over all queries,
\begin{equation}
C(q,g)=2C_{\mathrm{cls}}(q,g)+5C_{\ell_1}(q,g)+2C_{\mathrm{GIoU}}(q,g).
\label{eq:appendix_fragmentation_cost}
\end{equation}
The five queries with the lowest costs form the candidate set $\mathcal{C}_g^5$. Predicted and ground-truth boxes use normalized $(c_x,c_y,w,h)$ coordinates. For a candidate $q$ and object $g$, the center and scale discrepancies are
\begin{align}
e_{\mathrm{ctr}}(q,g)
&=\sqrt{\left(\frac{c_x^q-c_x^g}{w^g}\right)^2
+\left(\frac{c_y^q-c_y^g}{h^g}\right)^2},\nonumber\\
e_{\mathrm{scl}}(q,g)
&=\left|\log\frac{w^q}{w^g}\right|
+\left|\log\frac{h^q}{h^g}\right|.
\label{eq:appendix_fragmentation_box_errors}
\end{align}
Within $\mathcal{C}_g^5$, $q_{\mathrm{cls}}$, $q_{\mathrm{ctr}}$, $q_{\mathrm{scl}}$, and $q_{\mathrm{IoU}}$ denote the queries with the highest sigmoid score for the ground-truth class, smallest $e_{\mathrm{ctr}}$, smallest $e_{\mathrm{scl}}$, and highest box IoU, respectively. The owner $q_{\mathrm{own}}$ is obtained independently from the full O2O Hungarian assignment.

\begin{table*}[hbtp]
\centering
\small
\setlength{\tabcolsep}{5.2pt}
\begin{tabular}{@{}lrrrrrrr@{}}
\toprule
& & \multicolumn{3}{c}{Four-dimensional}
& \multicolumn{3}{c}{Localization-only} \\
\cmidrule(lr){3-5}\cmidrule(l){6-8}
\textbf{Group} & \textbf{\#GT}
& $\mathbf{A}_g$ (\%) & $\mathbf{D}_g$ & $\mathbf{R}_g$ (\%)
& $\mathbf{A}_g^{\mathrm{loc}}$ (\%) & $\mathbf{D}_g^{\mathrm{loc}}$ & $\mathbf{R}_g^{\mathrm{loc}}$ (\%) \\
\midrule
All & 99,481 & 8.48 & 2.33 & 56.02 & 53.36 & 1.51 & 69.49 \\
\midrule
Large  & 34,834 & 13.37 & 2.18 & 61.08 & 62.64 & 1.41 & 74.32 \\
Medium & 41,774 & 7.36  & 2.33 & 56.84 & 53.88 & 1.51 & 71.54 \\
Small  & 22,873 & 3.08  & 2.53 & 46.82 & 38.29 & 1.69 & 58.37 \\
\midrule
Owner IoU $\ge 0.5$ & 95,386 & 8.83 & 2.31 & 57.37 & 54.46 & 1.50 & 71.14 \\
Owner IoU $< 0.5$   & 4,095  & 0.27 & 2.73 & 24.59 & 27.79 & 1.86 & 31.08 \\
\bottomrule
\end{tabular}
\caption{Query-evidence fragmentation on CrowdHuman. $A$ is winner agreement, $D$ is the number of distinct winners, and $R$ is matched-owner coverage. The localization-only variants retain the original full-cost candidate set and owner but compute $A/D/R$ over the center, scale, and IoU winners only. Lower $A$ and $R$ and higher $D$ indicate stronger fragmentation.}
\label{tab:crowdhuman_fragmentation}
\end{table*}

For each object, the three quantities visualized in Fig.~\ref{fig:knowledge_fragmentation} are
\begin{align}
A_g &= \mathbb{I}[q_{\mathrm{cls}}=q_{\mathrm{ctr}}=q_{\mathrm{scl}}=q_{\mathrm{IoU}}],\nonumber\\
D_g &= \left|\{q_{\mathrm{cls}},q_{\mathrm{ctr}},q_{\mathrm{scl}},q_{\mathrm{IoU}}\}\right|,\nonumber\\
R_g &= \frac{1}{4}\sum_{d\in\{\mathrm{cls},\mathrm{ctr},\mathrm{scl},\mathrm{IoU}\}}
\mathbb{I}[q_{\mathrm{own}}=q_d].
\label{eq:appendix_fragmentation_metrics}
\end{align}
Panel (a) reports the group-wise mean of $A_g$ as a percentage, panel (b) reports the mean number $D_g\in[1,4]$ of distinct dimension-wise winners, and panel (c) reports the mean owner coverage $R_g$ as a percentage. Within each dataset, all three panels use one dataset-specific frozen baseline checkpoint; no averaging across checkpoints or method variants is performed. The side-by-side bars therefore test whether the diagnostic pattern recurs in a second dataset/model setting, rather than treating the dataset as the only changed factor in a controlled comparison. Because candidate ranking and the four dimension-wise winners use ground-truth information, these statistics are used only to diagnose O2O supervision and are not inputs to the prediction-aware graph at training or inference time.

\subsection{Cross-Dataset Fragmentation on CrowdHuman}

To test whether query-evidence fragmentation is specific to the evaluated COCO setting, we apply the same frozen, ground-truth-aware top-5 diagnostic to a 120-epoch DEIM baseline with an HGNetV2-B4 backbone on the complete CrowdHuman validation set. The analysis uses the checkpoint's EMA weights, deterministic validation transforms, and the full O2O Hungarian assignment. Using post-resize box areas, instances below $32^2$, from $32^2$ to $96^2$, and at least $96^2$ pixels$^2$ are grouped as small, medium, and large, respectively. The diagnostic covers 4,370 images and 99,481 full-body person instances; no training, gradient computation, or parameter update is performed.

CrowdHuman has only one foreground class, so counting $q_{\mathrm{cls}}$ as a fourth winner may influence the four-dimensional metric. As a control, we retain the original full-cost top-5 candidate set and Hungarian owner but omit $q_{\mathrm{cls}}$ from the reported winner set, defining
\begin{align}
A_g^{\mathrm{loc}}
&=\mathbb{I}[q_{\mathrm{ctr}}=q_{\mathrm{scl}}=q_{\mathrm{IoU}}],\nonumber\\
D_g^{\mathrm{loc}}
&=\left|\{q_{\mathrm{ctr}},q_{\mathrm{scl}},q_{\mathrm{IoU}}\}\right|,\nonumber\\
R_g^{\mathrm{loc}}
&=\frac{1}{3}\sum_{d\in\{\mathrm{ctr},\mathrm{scl},\mathrm{IoU}\}}
\mathbb{I}[q_{\mathrm{own}}=q_d].
\label{eq:appendix_crowdhuman_localization_metrics}
\end{align}

Tab.~\ref{tab:crowdhuman_fragmentation} shows that dispersion remains after the class winner is omitted from the reported metric: only $53.36\%$ of all instances share one center/scale/IoU-best query. Fragmentation is strongest for small persons: their localization-only agreement falls to $38.29\%$, while the matched owner covers only $58.37\%$ of the three localization winners. This control is not a localization-only rematching experiment, because candidate selection and owner assignment still use the detector's original full Hungarian cost. Poorly localized owners are also associated with stronger fragmentation than owners with IoU at least $0.5$, but this split is descriptive and is not independent of owner coverage because $R_g$ contains the owner--IoU-winner indicator. We therefore do not use it as independent evidence that fragmentation causes localization errors or AP changes.

\subsection{Scale-Wise Precision--Recall Profiles on CrowdHuman}
\label{app:crowdhuman_pr}

\begin{figure}[!t]
\centering
\includegraphics[width=0.8\columnwidth]{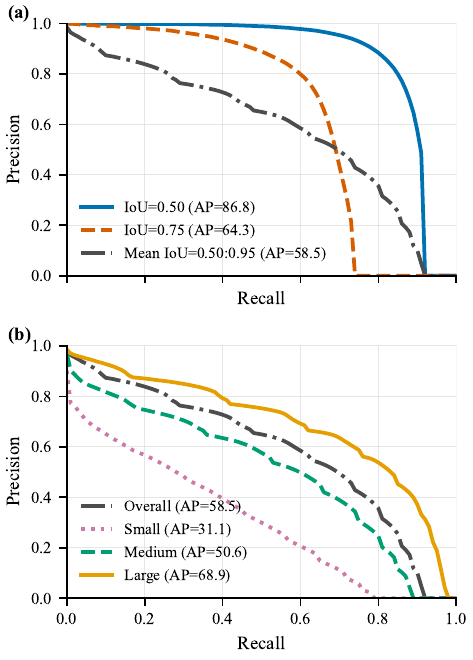}
\caption{Precision--recall profiles of the 100-epoch BS-O2G checkpoint on CrowdHuman. Panel (a) shows all-scale curves at IoU $0.50$, IoU $0.75$, and after averaging over IoU $0.50{:}0.95$; panel (b) shows the IoU-averaged curves by COCO object scale. All curves use 101 recall thresholds and at most 100 detections per image.}
\label{fig:crowdhuman_pr_100e}
\end{figure}

To examine why scale-specific AP can differ substantially from overall AP, Fig.~\ref{fig:crowdhuman_pr_100e} reports the interpolated precision--recall profiles saved by the COCO-style evaluator for the 100-epoch BS-O2G checkpoint on CrowdHuman. This is a descriptive single-checkpoint analysis rather than a comparison with DEIM or the 120-epoch best checkpoint reported in Tab.~\ref{tab:crowdhuman_deim_reference}.

This checkpoint obtains $58.5$ AP, $86.8$ AP$_{50}$, $64.3$ AP$_{75}$, $31.1$ AP$_S$, $50.6$ AP$_M$, and $68.9$ AP$_L$. The large-object curve remains above the overall and small/medium curves across most of the recall range, whereas the small-object curve loses precision earlier. The resulting scale concentration explains how AP$_L$ can be much higher than overall AP: overall AP is recomputed over all object scales under a unified confidence ranking, rather than formed as a linear average of AP$_S$, AP$_M$, and AP$_L$. Since Fig.~\ref{fig:crowdhuman_pr_100e} contains no baseline curve, it does not by itself attribute the $13.6$-point AP$_L$ difference in Tab.~\ref{tab:crowdhuman_deim_reference} to BS-O2G or identify its cause.

\subsection{Additional Efficiency Measurements}
\label{app:additional_efficiency}

We complement the main-text complexity and batch-1 latency results with batch-8 throughput, inference memory, and training-cost measurements in Tab.~\ref{tab:additional_efficiency}. All three variants are profiled on the same GPU and software environment with $640\times640$ inputs and AMP. Each inference repetition contains 30 warm-up iterations followed by 100 timed iterations. Training measurements use batch size 1, with 5 warm-up and 20 timed forward-and-backward iterations. We repeat each measurement three times, synchronize the GPU around each timed region, and report CUDA peak allocated memory. The training profile computes the original detection losses and gradients but deliberately excludes the optimizer update.

\begin{table*}[t]
\centering
\small
\setlength{\tabcolsep}{5pt}
\begin{tabular}{@{}lccccc@{}}
\toprule
\textbf{Method}
& \textbf{Throughput$_{b=8}$ (img/s)}
& \textbf{Infer. Mem.$_{b=1}$ (MiB)}
& \textbf{Infer. Mem.$_{b=8}$ (MiB)}
& \textbf{Fwd.+Bwd.$_{b=1}$ (ms)}
& \textbf{Train Mem.$_{b=1}$ (MiB)} \\
\midrule
DEIM baseline
& $324.2{\pm}4.3$
& 588.1
& 853.9
& $202.5{\pm}4.2$
& $1097.5{\pm}34.8$ \\
\quad + O2G
& $300.3{\pm}4.1$
& 589.8
& 855.6
& $203.5{\pm}1.6$
& $1068.2{\pm}84.0$ \\
\textbf{\quad + BS-O2G}
& $298.5{\pm}3.1$
& 590.4
& 856.2
& $228.3{\pm}30.3$
& $1254.6{\pm}164.9$ \\
\bottomrule
\end{tabular}
\caption{Additional efficiency measurements with a ResNet-50 backbone. Values are mean $\pm$ standard deviation over three repetitions when the measured variation is nonzero. Inference throughput uses batch size 8; forward-and-backward time and peak training memory use batch size 1 and exclude the optimizer update.}
\label{tab:additional_efficiency}
\end{table*}

At batch size 8, BS-O2G processes $298.5$ images/s, a $7.9\%$ reduction from the baseline and only $0.6\%$ below O2G. Its peak inference memory differs from the baseline by only $2.3$ MiB for both batch sizes. The complete BS-O2G model shows approximately $12.7\%$ longer forward-and-backward time and $14.3\%$ higher peak training memory than the baseline in this profile. However, the comparatively large deviations of the training measurements, especially for BS-O2G, indicate sensitivity to shared-device load; these values characterize the observed overhead rather than a hardware-independent constant. The lower mean training memory of O2G relative to the baseline is likewise within this measurement variability and should not be interpreted as a systematic memory saving.

\section{Additional Mechanism Analysis}
\label{app:mechanism_analysis}
\label{app:relation_diagnostics}

\subsection{Role of BS}
\label{app:role_bs}

The staged ablation in Tab.~\ref{tab:training_inference_ablation} clarifies that BS is not the dominant source of the overall AP improvement. O2G and the persistent basis $U$ increase the baseline from 53.0 to 53.4 AP, while adding BS further raises the result to 53.5 AP. Unlike O2G, BS adds no forward prediction path and leaves the forward values unchanged. With $\lambda_{\mathrm B}=0.02$, it conservatively combines the local basis gradient with a graph-routed component, refining how the existing O2O optimization signals accumulate in $U$. The lower results of BS without O2G (52.7 AP) and random BS routing (52.8 AP) further indicate that this contribution depends on its coupling with O2G and the learned graph. We therefore interpret BS as a modest optimization complement within the integrated framework rather than the primary source of its total gain.

\subsection{Persistent Query-Basis Response Analysis}
\label{app:basis_response_analysis}

\subsubsection{Cross-Image Class-Wise Response Consistency}
\label{app:u_response_stability}

Using a fixed BS-O2G checkpoint with O2G disabled, we evaluate 1,000 COCO validation images and isolate each basis vector's marginal class-logit effect as $\Delta_U(c)=z_c(Q+U)-z_c(Q)$. For each of the 300 basis vectors, we average this effect separately over two disjoint image halves and compute the Spearman correlation between the resulting 80-class response profiles.

The distribution in Fig.~\ref{fig:u_response_stability} is predominantly positive, with a median correlation of $0.417$. This indicates that $U$ contains repeatable, basis-vector-specific class-response patterns across images. These patterns are soft and distributed rather than one-basis-per-class prototypes, and this fixed-checkpoint diagnostic does not by itself establish that the response structure causes the AP gain.

\begin{table}[htbp]
\centering
\small
\begin{tabular}{@{}lcccc@{}}
\toprule
\textbf{Trial} & \textbf{Follow-$U$} & \textbf{Stay-slot} & \textbf{Gap} & \textbf{95\% CI} \\
\midrule
1 & 0.5114 & 0.3028 & +0.2086 & [0.1231, 0.2934] \\
2 & 0.5326 & 0.3069 & +0.2257 & [0.1401, 0.3163] \\
3 & 0.5143 & 0.2647 & +0.2497 & [0.1600, 0.3340] \\
\bottomrule
\end{tabular}
\caption{No-fixed-point $U$-swap control for query-slot bias. The gap is Follow-$U$ minus Stay-slot.}
\label{tab:u_swap_slot_control}
\end{table}

\subsubsection{$U$-Swap Control for Query-Slot Bias}
\label{app:u_swap_slot_control}

We test whether the cross-image response consistency can be explained solely by each basis vector remaining at a fixed query slot. In each of three independent no-fixed-point permutations, every $U$ basis vector is moved to a different slot while the checkpoint is fixed and O2G is disabled. We then compare the post-swap response profile with (i) the original profile of the moved basis vector (Follow-$U$) and (ii) the original profile at the destination slot (Stay-slot).

All three gaps are positive and their bootstrap confidence intervals exclude zero. The response therefore follows the moved basis vector more strongly than it remains at the original query position, showing that the cross-image consistency cannot be attributed solely to a fixed query-slot bias. Because this is a fixed-checkpoint intervention rather than a retrained ablation, it diagnoses response identity but does not by itself establish an AP gain.

\subsection{Cross-Image, Different-Owner U Reuse}
\label{app:cross_image_u_reuse}

As a stricter supplement to the parameter-access statistics in the main text, we compare Top-8 destination IDs between detections of the same class but from different images and different owner queries. The analysis covers 77 categories and 1,026,639 sampled detection pairs.

\begin{table}[H]
\centering
\small
\begin{tabular}{lc}
\toprule
Routing rule & Mean overlap \\
\midrule
Q-only & 0.674 \\
U-only & 5.172 \\
Shared-U (Q+U) & \textbf{2.914} \\
Query-private-U & 0.000 \\
Shuffled destinations & 0.214 \\
\bottomrule
\end{tabular}
\caption{Mean Top-8 destination-ID overlap for same-class detections from different images and different owner queries.}
\label{tab:cross_image_u_reuse}
\end{table}

Shared Q+U routing retains $2.914$ common destinations, compared with $0$ under query-private access and $0.214$ after destination shuffling. This supports cross-image reuse at the routing level. Because the query-private result is a fixed-checkpoint access-rule counterfactual rather than a retrained model, this diagnostic does not establish an AP gain caused by shared access.

\subsection{Controls for Semantic Gradient Routing}
\label{app:semantic_routing_controls}

The main text directly compares learned and random $U$ routing. We further test whether the related-pair advantage can be explained by category frequency or commonly used $U$ basis vectors.

\noindent\textbf{Category-frequency control.}
Let $n_a$ and $n_b$ be the numbers of matched objects for categories $a$ and $b$. For every pair, we regress each routing metric on the pair's mean log-frequency, $[\log(1+n_a)+\log(1+n_b)]/2$, and its absolute log-frequency difference, $|\log(1+n_a)-\log(1+n_b)|$. We then compare the residual metric between related and unrelated pairs. The controlled gap in Tab.~\ref{tab:semantic_routing_controls} is the difference between these two residual means. Significance is evaluated using 5,000 permutations of the COCO super-category labels that preserve group sizes, followed by Benjamini--Hochberg correction over 75 tests.

\noindent\textbf{Random-$U$ control.}
Random $U$ routing keeps the matched-owner gradients and routing weights unchanged but randomly reassigns their destination basis-vector IDs. For every category pair and metric, we compute the learned-routing score minus its random-$U$ score and apply the same frequency control. This removes alignment caused by general use of high-frequency basis-vector destinations rather than the learned destination structure.

The total-gradient result also persists in the successful-detection cohort. Under learned $U$ routing, related and unrelated pairs yield $0.0579$ and $0.0389$, with a controlled gap of $0.0197$ ($q=0.0004$). After subtracting random $U$ routing, the corresponding values are $0.0437$ and $0.0274$, with a controlled gap of $0.0170$ ($q=0.0009$).

\begin{table}[htbp]
\centering
\footnotesize
\setlength{\tabcolsep}{2.2pt}
\begin{tabularx}{\columnwidth}{@{}>{\raggedright\arraybackslash}Xcccc@{}}
\toprule
\textbf{Routing / metric} & \textbf{Related} & \textbf{Unrel.} & \textbf{Ctrl. gap} & \textbf{BH $q$} \\
\midrule
Learned: Jaccard & .5754 & .5275 & +.0388 & .0008 \\
Learned: total-grad. cosine & .0487 & .0390 & +.0108 & .0112 \\
Learned: shared-basis cosine & .1426 & .1074 & +.0367 & .0008 \\
\midrule
Learned--random: Jaccard & -.0299 & -.0684 & +.0408 & .0008 \\
Learned--random: total-grad. cosine & .0309 & .0235 & +.0082 & .0222 \\
Learned--random: shared-basis cosine & .0668 & .0442 & +.0219 & .0008 \\
\bottomrule
\end{tabularx}
\caption{Semantic-routing controls over all 3,160 unordered COCO category pairs. Related pairs share an official COCO super-category.}
\label{tab:semantic_routing_controls}
\end{table}

\subsection{Top-$K$ Neighborhood Stability}
\label{app:topk_jaccard}

We evaluate whether prediction-aware query neighborhoods remain stable across decoder layers. For each validation image, we record the features $h_i^{(\ell)}$, boxes $b_i^{(\ell)}$, and logits $z_i^{(\ell)}$ of all $N=300$ query indices at each of the six decoder layers, with $p_i^{(\ell)}=\sigma(z_i^{(\ell)})$. At layer $\ell$, the diagnostic affinity between two distinct queries is
\begin{equation}
\begin{aligned}
a_{ij}^{(\ell)}
&=\operatorname{cos}(h_i^{(\ell)},h_j^{(\ell)})
+\operatorname{IoU}(b_i^{(\ell)},b_j^{(\ell)})\\
&\quad+\operatorname{cos}(p_i^{(\ell)},p_j^{(\ell)}).
\end{aligned}
\label{eq:appendix_affinity}
\end{equation}
After excluding the self-edge, we retain the $K=8$ highest-affinity neighbors of query $i$ as $\mathcal{N}_i^{(\ell)}$.

For each query and each adjacent layer pair, neighborhood stability is measured by
\begin{equation}
J_i^{(\ell)}=
\frac{|\mathcal{N}_i^{(\ell)}\cap\mathcal{N}_i^{(\ell+1)}|}
{|\mathcal{N}_i^{(\ell)}\cup\mathcal{N}_i^{(\ell+1)}|}.
\label{eq:appendix_jaccard}
\end{equation}
We first average $J_i^{(\ell)}$ over the 300 queries and the five adjacent layer pairs within each image, and then average the image-level values over all 5,000 COCO validation images. The resulting mean top-$K$ Jaccard is $0.6429$, indicating that the prediction-aware neighborhoods are not confined to the final decoder layer but remain substantially consistent during iterative decoding.

\subsection{Neighbor Intervention}
\label{app:neighbor_intervention}

We further test whether the diagnostic neighborhoods have a measurable effect on the anchor prediction. On a 512-image subset, we select up to two high-confidence anchor queries per image from true-positive owners, poor-IoU matched queries, duplicate false positives, and localization false positives. For each anchor, we form three size-8 neighbor sets using the final-layer affinity: the highest-affinity queries (Top), randomly sampled non-self queries (Random), and the lowest-affinity queries (Low). At the input of the last decoder layer, the features of the selected neighbors are replaced by the image-wide mean query feature, while the anchor feature is left unchanged. We then rerun inference and measure the anchor's class change by the Jensen--Shannon divergence between the softmax-normalized logits and its box change by the $L_1$ distance over normalized box coordinates.

\begin{table}[H]
\centering
\small
\begin{tabular}{lcc}
\toprule
Neighbors & Class JS $\uparrow$ & Box $L_1$ $\uparrow$ \\
\midrule
Top & $2.5138\!\times\!10^{-5}$ & $2.2855\!\times\!10^{-8}$ \\
Random & $4.6265\!\times\!10^{-7}$ & $3.8282\!\times\!10^{-9}$ \\
Low & $7.8142\!\times\!10^{-6}$ & $1.2532\!\times\!10^{-8}$ \\
\bottomrule
\end{tabular}
\caption{Effect of replacing different neighbor sets.}
\label{tab:intervention}
\end{table}

Replacing top neighbors changes the class distribution about $54.3\times$ more than replacing random neighbors. However, the absolute box changes are small. We therefore treat this experiment as supporting evidence that high-affinity neighbors have a functional influence on class prediction, rather than as a strong causal claim about localization.

\section{Additional Qualitative Results}
\label{app:qualitative_results}

\subsection{Additional Paired Recovery Cases}
\label{app:paired_attention_cases}

\begin{figure*}[t]
\centering
\includegraphics[width=0.98\textwidth]{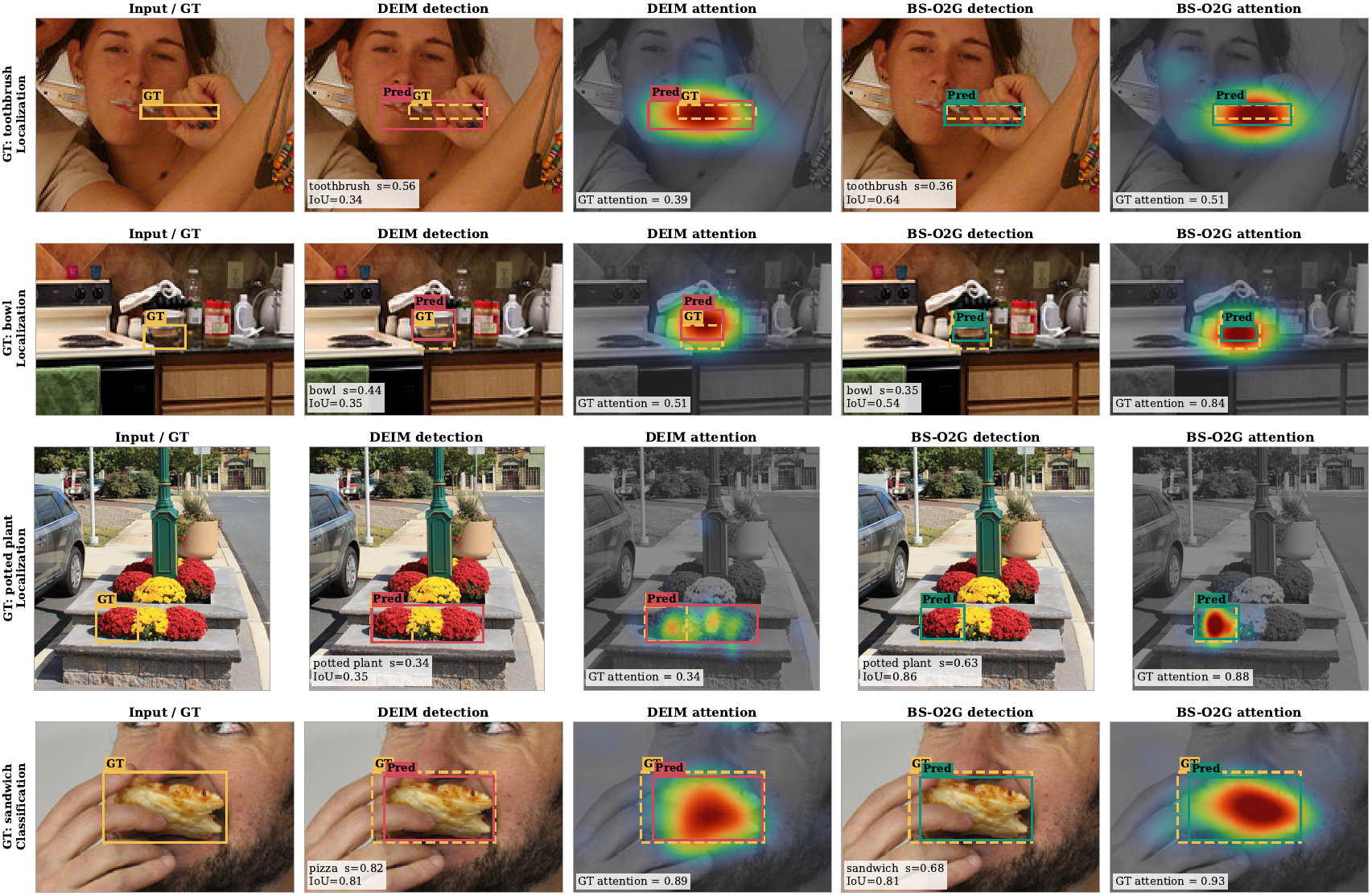}
\caption{Additional paired recoveries, continued from Fig.~\ref{fig:paired_attention_main}. From top to bottom: toothbrush, bowl, and potted-plant localization recoveries, followed by a sandwich classification recovery. Columns, boxes, attention heatmaps, and within-row color scaling follow the same conventions as the main-text figure.}
\label{fig:paired_attention_appendix_a}
\end{figure*}

\begin{figure*}[t]
\centering
\includegraphics[width=0.98\textwidth]{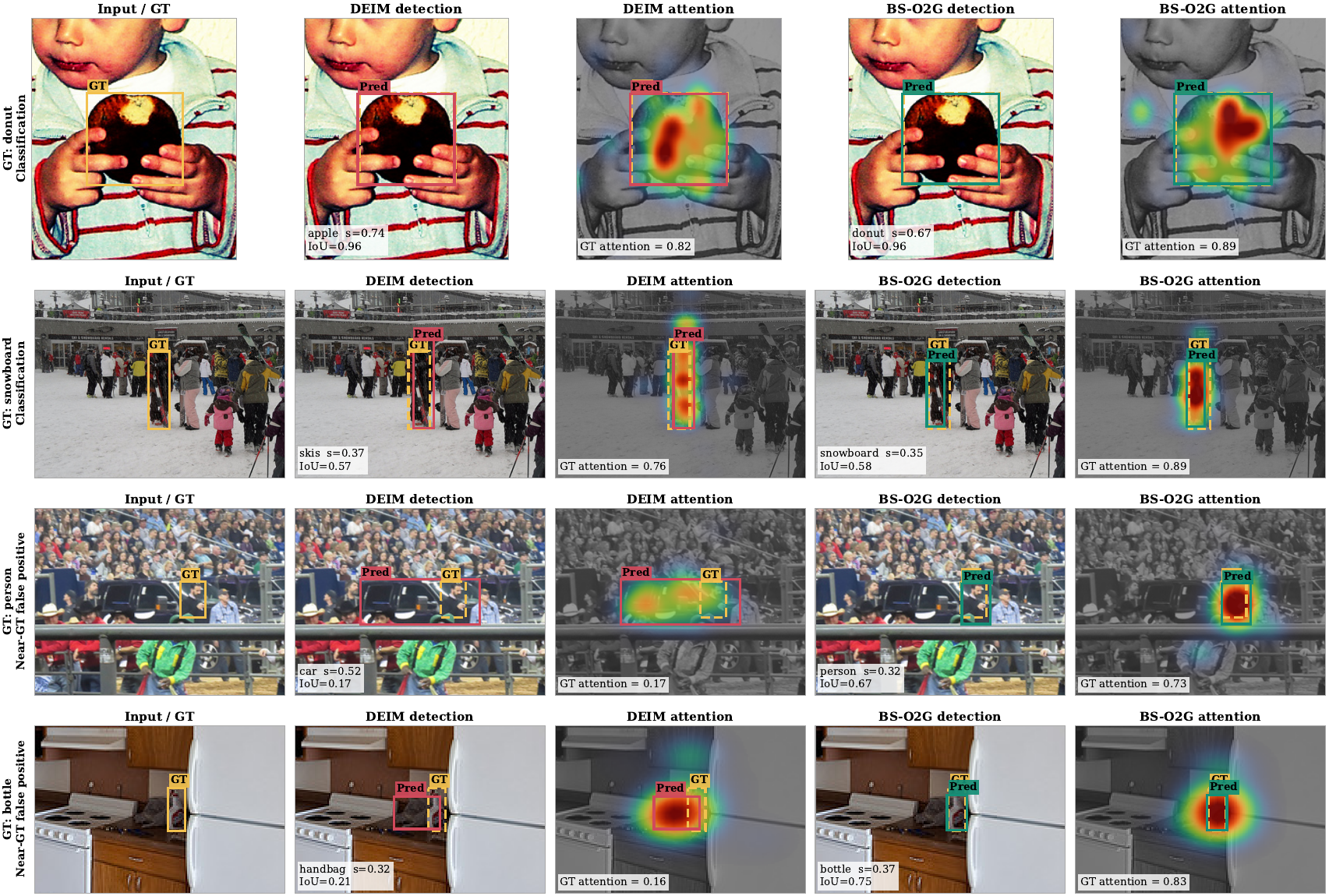}
\caption{Additional paired recoveries, continued from Fig.~\ref{fig:paired_attention_appendix_a}. From top to bottom: donut and snowboard classification recoveries, followed by person and bottle near-GT false-positive recoveries.}
\label{fig:paired_attention_appendix_b}
\end{figure*}

Figs.~\ref{fig:paired_attention_appendix_a} and~\ref{fig:paired_attention_appendix_b} present the eight cases not selected for Fig.~\ref{fig:paired_attention_main}. Candidate targets are frozen from the full-validation DEIM-non-TP to BS-O2G-TP transition table. For each model, boxes, scores, and last-decoder deformable cross-attention are recomputed in the same AMP forward pass from the actual score-thresholded final detection query. The examples are selected within localization, classification, and near-GT false-positive strata using fixed quantiles and a positive GT-attention-gain filter. They therefore illustrate representative recovered detections, but are not an unbiased estimate of COCO-wide behavior.

\end{document}